\RequirePackage{fix-cm}
\documentclass[twocolumn, natbib]{svjour3}          
\smartqed  
\usepackage{times}
\usepackage{epsfig}
\usepackage{graphicx}
\usepackage[cmex10]{amsmath}
\usepackage{amssymb}
\usepackage{tabulary,color}
\usepackage{tabularx}
\usepackage{multirow}
\usepackage{booktabs, siunitx, dcolumn}
\usepackage{colortbl}
\usepackage{pifont}
\usepackage{dsfont}
\usepackage{enumitem} 
\usepackage[para]{threeparttable} 
\usepackage[export]{adjustbox}
\usepackage[percent]{overpic}
\usepackage{array} 
\usepackage{url} 
\usepackage{fixltx2e} 
\usepackage{algorithm}     
\usepackage{algorithmic}     
\usepackage[caption=false,font=footnotesize]{subfig}
\usepackage{xcolor}

\usepackage{natbib}
\setcitestyle{authoryear,open={(},close={)}} 

\definecolor{link}{RGB}{75, 166, 154}
\usepackage[pagebackref=true,breaklinks=true,colorlinks,citecolor=blue,bookmarks=false]{hyperref} 

\usepackage[toc,page]{appendix} 

\usepackage{url}

\usepackage{marvosym}

\usepackage{xspace}
\usepackage{ragged2e}
\usepackage[labelfont=bf, font=footnotesize, justification=justified, format=plain]{caption}

\usepackage{makecell} 
\newcommand{\OM}{GMCNet}

\journalname{International Journal of Computer Vision}
\begin{document}\sloppy

\title{Robust Partial-to-Partial Point Cloud Registration in a Full Range
}

\author{Liang Pan$^{1}$         \and
        Zhongang Cai$^{1,2,3}$    \and
        Ziwei Liu$^{1\href{mailto:ziwei.liu@ntu.edu.sg}{\textrm{\Letter}}}$
}


\institute{
$^{1}$ S-Lab, Nanyang Technological University, Singapore \\
$^{2}$ Sensetime Research\\
$^{3}$ Shanghai AI Lab.\\
Corresponding author: Ziwei Liu \at ziwei.liu@ntu.edu.sg
}

\date{Received: date / Accepted: date}

\maketitle

\begin{abstract}
\justifying
Point cloud registration for 3D objects is a challenging task due to sparse and noisy measurements, incomplete observations and large transformations.
In this work, we propose \textbf{G}raph \textbf{M}atching \textbf{C}onsensus \textbf{Net}work (\textbf{\OM{}}), which estimates pose-invariant correspondences for full-range Partial-to-Partial point cloud Registration (PPR) in the object-level registration scenario.
To encode robust point descriptors, \textbf{1)} we first comprehensively investigate transformation-robustness and noise-resilience of various geometric features.
\textbf{2)} Then, we employ a novel {T}ransformation-robust {P}oint {T}ransformer (\textbf{TPT}) module to adaptively aggregate local features regarding the structural relations, which takes advantage from both handcrafted rotation-invariant ({\textit{RI}}) features and noise-resilient spatial coordinates.
\textbf{3)} Based on a synergy of hierarchical graph networks and graphical modeling, we propose the {H}ierarchical {G}raphical {M}odeling (\textbf{HGM}) architecture to encode robust descriptors consisting of 
i) a unary term learned from {\textit{RI}} features; 
and ii) multiple smoothness terms encoded from neighboring point relations at different scales through our TPT modules.
Moreover, we construct a challenging PPR dataset (\textbf{MVP-RG}) based on the recent MVP dataset that features high-quality scans.
Extensive experiments show that \OM{} outperforms previous state-of-the-art methods for PPR.
Notably, \OM{} encodes point descriptors for each point cloud individually without using cross-contextual information, or ground truth correspondences for training.
Our code and datasets are available at \href{https://github.com/paul007pl/GMCNet}{https://github.com/paul007pl/GMCNet}.
\end{abstract}

\begin{figure*}
    \centering
    \includegraphics[width=0.95\linewidth]{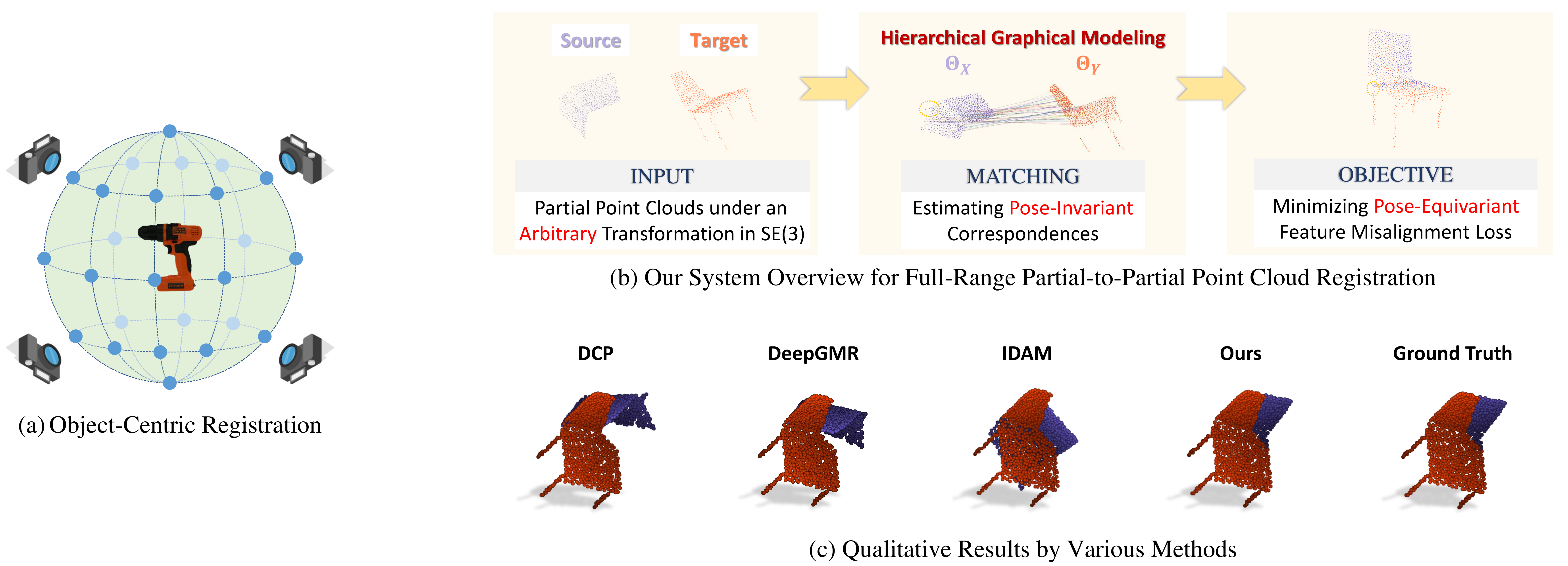}
    \caption{
    \textbf{(a) Object-Centric Registration Application.} 
    Object 6D-pose estimation is an example application that often requires to register partial point clouds under large transformations, which are observed from diverse viewpoints.
    \textbf{(b) System Overview.} 
    Our \OM{} learns robust feature descriptors for full-range Partial-to-Partial point cloud Registration (PPR). 
    \textbf{(c) Qualitative Comparison.} 
    Qualitative results for full-range PPR on ModelNet40~\citep{wu20153d} by various methods, including DCP~\citep{wang2019deep}, DeepGMR~\citep{yuan2020deepgmr}, IDAM~\citep{li2019iterative} and \OM{}.
    }
    \label{fig:teaser}
\end{figure*}
\section{Introduction}
\label{sec:introduction}


Point cloud registration (PCR) is a fundamental task for a wide range of applications such as localization~\citep{droeschel2018efficient, yang2018robust, ding2019deepmapping}, 3D reconstruction~\citep{wan2018robust, lu2019l3} and 6D object pose estimation~\citep{wong2017segicp, tian2020robust}, where a rigid transformation (\emph{i.e}, a 3D rotation and a 3D translation) is estimated to align a source point cloud to a target point cloud. 
Classic methods that leverage Iterative Closest Point algorithm require a good transformation initialization and usually prune to local minima.
With the advent of deep learning, researchers are encouraged to creating learning models for solving PCR. 

A series of learning-based PCR methods~\citep{wang2019deep, wang2019prnet, li2019iterative, yew2020rpm, fu2021robust} have been proposed. 
In the {object-centric} PCR scenario, Wang et. al.~\citep{wang2019deep, wang2019prnet} learn to encode point features for estimating a mapping between the source point cloud and the target from ModelNet40 dataset~\citep{wu20153d}.
Instead of directly taking 3D coordinates as inputs, recent PCR methods, RPMNet~\citep{yew2020rpm}, DeepGMR~\citep{yuan2020deepgmr} and IDAM~\citep{li2019iterative}, provide better registration results with the help of rotation-invariant ({\textit{RI}}) features.
However, most of them (except DeepGMR) are only validated under restricted source-target transformations (\emph{i.e.}, rotations in [0, 45{\textdegree}]).
According to our experiments, large transformations could undermine their training stability, and also largely increase registration error.
Although DeepGMR could handle PCR with large transformations for complete point clouds, 
partial point clouds significantly degrade its registration performance.
In practice,
unconstrained viewpoints give rise to large transformations (\emph{e.g.}, Fig.~\ref{fig:teaser} (a) demonstrates arbitrary transformations in the SE(3)) along with noisy measurements and partial observations, which are the key challenges for PCR.
In this work, we take one step further and focus on partial-to-partial point cloud registration (PPR) in the \textit{full-range}~\footnote{In this paper, full-range transformation and global transformation are used interchangeably, both refer to arbitrary transformations in SE(3).}, which are more relevant to real-world applications.

Encoding robust and discriminative point feature descriptors against large transformations and noisy measurements could be the major difficulty in estimating correct correspondences for global registration.
Motivated by the success of using {\textit{RI}} features for unconstrained arbitrary rotations in 3D object classification~\citep{drost2010model, chen2019clusternet, li2021rotation} and 3D molecule property prediction~\citep{fuchs2020se, satorras2021n}, we find \textit{RI} features promising, given carefully designed network architectures, to outperform existing methods~\citep{yew2020rpm, yuan2020deepgmr, li2019iterative} for full-range PPR.
To better understand the advantages and limitations of conventional coordinate encoding and various handcrafted \textit{RI} features (including \textit{RRI}, \textit{FPFH} and \textit{PPF}), we start by analyzing their robustness against arbitrary transformations and random noises.
On the one hand, we observe that {\textit{RI}} features between different partial point clouds with arbitrary SE(3) transformations could be no longer invariant, but they are still more transformation-robust than coordinate encoding.
On the other hand, \textit{RI} features are much more sensitive to noise than coordinate features.
Despite that {\textit{RI}} features are inconsistent due to different noisy observations, {\textit{RI}} features with jitters augmentations could facilitate {\textit{RI}} classification~\citep{chen2019clusternet, li2021rotation}, and hence we speculate that they still describe {\textit{RI}} 3D geometric properties. 
In addition, global and relative spatial coordinates of point clouds show superior noise-resilient properties, which results in stable structural relations.
In view of this,
we are motivated to match robust distribution-level features learned from structured {\textit{RI}} features for global PPR.

Equipped with the findings from the preliminary study, we develop our approach to full-range PPR. We represent point cloud as a set of features with graph structures, and the correspondence estimation for PPR is reformulated as a maximum common subgraph prediction problem. Specifically, we propose a one-shot paradigm, {G}raph {M}atching {C}onsensus {Net}work (\textbf{\OM{}}), which estimates pose-invariant correspondences (Fig.~\ref{fig:teaser} (b)) using {\textit{RI}} features and multi-scale graph structures for full-range PPR. In order to encode large-scale distribution features, we enlarge the receptive fields by consecutively sub-sampling point clouds in a hierarchical architecture.
To tackle the uncertainty introduced in the sub-sampling process, \emph{e.g} Farthest Point Sampling (FPS)~\footnote{FPS usually initials the sampling with the first point.} that leads to inconsistent encoded features, we employ a novel {T}ransformation-robust {P}oint {T}ransformer (\textbf{TPT}) module, which adaptively aggregates neighboring features from sampled points with the help of both {\textit{RI}} feature graphs and noise-resilient spatial relations. Following the idea of graphical modeling, we further propose the {H}ierarchical {G}raphical {M}odeling (\textbf{HGM}) architecture that encodes robust descriptors for each point consisting of: 1) a unary term learned from {\textit{RI}} features; and 2) multiple smoothness terms to encourage spatially-smooth correspondences with multi-scale geometric distribution-level features.

Moreover, following the recent MVP dataset~\citep{pan2021variational}, we construct a challenging object-centric PPR benchmark dataset (dubbed \textbf{MVP-RG}) consisting of 7,600 high-quality point cloud pairs.
Compared with the standard object-centric benchmark dataset, ModelNet40~\citep{wu20153d}, partial point clouds in MVP-RG are generated by high-resolution virtual cameras from diverse viewpoints, which better imitates real observations than uniformly sampling used for ModelNet40.
Experimental results on ModelNet40 and MVP-RG dataset show that \OM{} outperforms previous state-of-the-art (SoTA) methods.
Qualitative results for full-range PPR on ModelNet40 by various methods are shown in Fig.~\ref{fig:teaser} (c), where \OM{} achieve accurate registration in spite of noisy and partial observation. We highlight that \OM{} employs a one-shot paradigm for PPR, and it encodes robust descriptors for each point cloud individually without requiring cross-contextual information. In other words, the processing of each point clouds is self-sufficient and does not require information of the other point cloud in the source-target pair. Besides, no ground truth correspondences are needed during training.

The key contributions are summarized as:
\vspace{-1.5mm}
\begin{itemize}
\item We comprehensively analyze transformation-robustness and noise-resilience of different geometric features, which paves the way to encode robust descriptors for global PPR.

\item Employing graphical modeling in the context of deep learning, we propose a novel hierarchical graph network (dubbed \OM{}) by employing novel TPT modules, which encodes robust geometric feature descriptors.

\item Extensive experimental results show that \OM{} achieves much better registration results than previous SoTA methods, especially for full-range PPR.

\item A challenging object-centric PPR benchmark dataset (MVP-RG) is constructed, which consists of 7,600 paired high-quality partial point clouds generated by using virtual cameras from various viewpoints.
\end{itemize}

\section{Related Work}
\subsection{Rotation-Invariant ({\textit{RI}}) Features}
Learning {\textit{RI}} features has been extensively studied.
By defining local reference frames (LRF), many descriptors~\citep{chen20073d,frome2004recognizing,johnson1999using,salti2014shot,tombari2010unique} accumulate measurements into histograms according to spatial coordinates, surface normals and curvatures.
However, it is difficult to achieve rotation-invariance by defining unambiguous LRFs. 
Without relying on LRFs, researchers designed handcrafted {\textit{RI}} features, such as \textit{PPF}~\citep{drost2010model}, PFH~\citep{rusu2008aligning} and \textit{FPFH}~\citep{rusu2009fast}.
With the help of deep learning, robust {\textit{RI}} descriptors can be encoded based on the handcrafted features for downstream tasks~\citep{deng2018ppfnet,deng2018ppf,zhao20193d,chen2019clusternet}.
For example, networks designed with \textit{RRI} features show impressive performance for classification~\citep{chen2019clusternet} and registration~\citep{yuan2020deepgmr,sun2020canonical}.
In addition, a few recent works~\citep{choy2019fully,gojcic2019perfect,bai2020d3feat} study on encoding geometric features with deep networks, while they mainly focus on learning 3D point descriptors from large-scale dense 3D point clouds.

\subsection{Deep Point Cloud Registration (PCR)}
Conventional PCR methods mainly use the ICP~\citep{besl1992method} and its variants~\citep{rusinkiewicz2001efficient,segal2009generalized,bouaziz2013sparse,pomerleau2015review}.
However, these methods~\citep{besl1992method,rusinkiewicz2001efficient,segal2009generalized,bouaziz2013sparse,pomerleau2015review} only can register point clouds under a small relative transformation, as they prune to local minima.
To guarantee a global optima, complicated optimization approaches, such as branch-and-bound~\citep{yang2013go}, convex relaxation~\citep{maron2016point}, line process~\citep{zhou2016fast} and semidefinite programming~\citep{yang2019polynomial}, are proposed.

In the object-centric scenario, many learning-based PCR methods are proposed recently.
Comparing to non-learning methods, many learning PCR methods could achieve better object-centric registration.
PointNetLK~\citep{aoki2019pointnetlk} and PCRNet~\citep{sarode2019pcrnet,sarode2019one} iteratively regress the rotation and translation using global features encoded by PointNet~\citep{qi2017pointnet}.
DCP~\citep{wang2019deep} and PRNet~\citep{wang2019prnet} generate the estimated transformation by predicting soft assignments between two point sets with encoded features by DGCNN~\citep{dgcnn}.
Other follow-up works~\citep{yew2020rpm, li2019iterative} iteratively resolve PPR with robust descriptors encoded by {\textit{RI}} features.
Nonetheless, most existing methods~\citep{aoki2019pointnetlk,sarode2019pcrnet,sarode2019one,wang2019deep,wang2019prnet, yew2020rpm,li2019iterative,li2020deterministic,jiang2021sampling,bauer2021reagent,xu2021omnet} focus on local PCR (\emph{i.e}, rotations in [0, 45\textdegree]), while overlooking global PCR.
Additionally, they prefer to learn point descriptors without using hierarchies for sparse object-centric PPR.
DeepGMR~\citep{yuan2020deepgmr} addresses full-range PCR for complete point clouds by using pose-invariant GMM latent correspondences. 
However, different partial point clouds mostly do not have the same GMM variables, which makes it inapplicable for PPR.
\begin{figure*}
    \centering
    \includegraphics[width=1.0\linewidth]{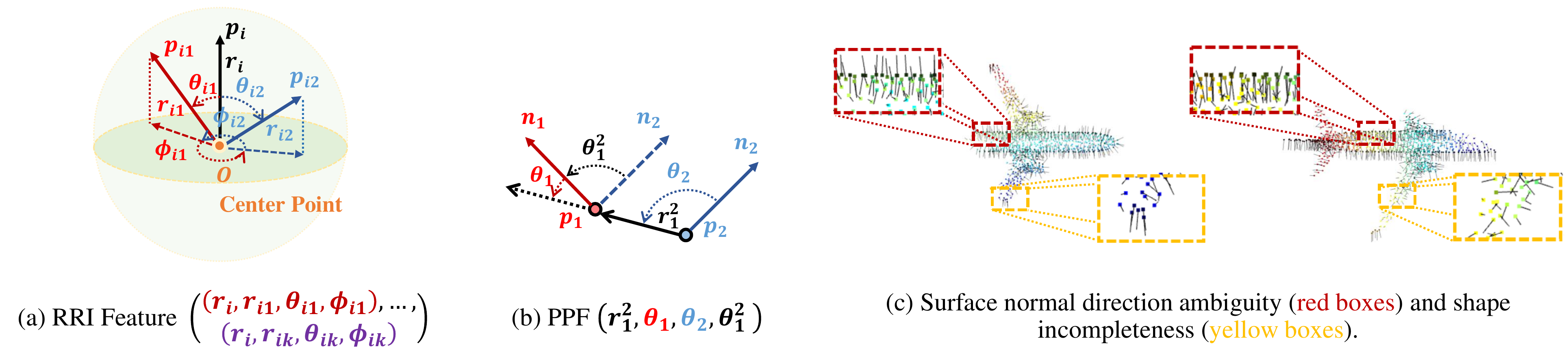}
    \caption{
    \textbf{Handcrafted Rotation-Invariant Features.}
    \textbf{(a)} \textit{RRI}~\citep{chen2019clusternet} 
    $r_i$ denotes the distance between $p_i$ and $O$ (e.g. $\|p_i\|_2$).
    $r_{ik}$ denotes the distance between the $k^{th}$ nearest neighbor of $p_i$ and $O$.  
    $\theta$ and $\phi$ are angles between $p_i$ and its neighboring points.  
    \textbf{(b)} \textit{PPF}~\citep{drost2010model} $r_1^2$ is the distance $\|p_1 - p_2\|_2$.
    $\theta_1$ and $\theta_2$ are $\angle (\protect\overrightarrow{p_1p_2}, \protect\overrightarrow{n_1})$ and $\angle (\protect\overrightarrow{p_1p_2}, \protect\overrightarrow{n_2})$, respectively.
    $\theta_1^2$ is $\angle (\protect\overrightarrow{n_1}, \protect\overrightarrow{n_2})$, the angle between $\protect\overrightarrow{n_1}$ and $\protect\overrightarrow{n_2}$.
    \textbf{(c)} 
    {Inconsistent surface normal estimation usually consists of 1. surface normal direction ambiguity (({\color[RGB]{192,0,0}{red boxes}})); 2. shape incompleteness ({\color[RGB]{255,192,0}{yellow boxes}}); and 3. inconsistent measurements that often caused by different 3D scans or noisy measurements.}
    }
    \label{fig:rot_feature}
\end{figure*}
\section{Problem Analysis} \label{sec:pa}
Following previous object-centric registration methods, in this work, we focus on small-scale object-centric point cloud registration.
In this section, we first formulate the Partial-to-Partial point cloud Registration (PPR), and then introduce various pose-invariant point features.
Finally, we analyze the transformation-robustness and noise-resilience of different handcrafted features.

\subsection{Partial-to-Partial Point Cloud Registration (PPR)}
Given a source point cloud $\mathcal{X}=\{x_1, ..., x_N\} \subset \mathbb{R}^3$ and a target point cloud $\mathcal{Y}=\{y_1, ..., y_M\} \subset \mathbb{R}^3$, PCR targets at aligning $\mathcal{X}$ to $\mathcal{Y}$, \emph{i.e.} mapping each $x_i$ to $y_{m(x_i)}$, by finding a rigid transformation $\mathbf{T}_{\mathcal{XY}} = [\mathbf{R}_{\mathcal{XY}},\: \mathbf{t}_{\mathcal{XY}}]$ comprising a rotation $\mathbf{R}_{\mathcal{XY}}\in$ SO(3) and a translation $\mathbf{t}_{\mathcal{XY}}\in \mathbb{R}^3$.
If $\mathcal{X}$ and $\mathcal{Y}$ are two different \textit{partial} point clouds with overlapped areas, it becomes the more challenging PPR problem, as more outliers are introduced by their different partial observations.
Previous methods~\citep{wang2019prnet, yew2020rpm} address the registration problem by alternating between two steps: 
1) predicting a mapping $m$ between $\mathcal{X}$ and $\mathcal{Y}$; 
2) estimating current optimal rotation $\mathbf{R}_{\mathcal{XY}}$ and translation $\mathbf{t}_{\mathcal{XY}}$ with the mapped inliers $\mathcal{X}_{in} \subset \mathcal{X}$ and $\mathcal{Y}_{in} \subset \mathcal{Y}$.
Given a predicted mapping $m$, the optimal $\mathbf{R}_{\mathcal{XY}} = \mathbf{VU}^\top$ and $\mathbf{t}_{\mathcal{XY}}=-\mathbf{R}_{\mathcal{XY}}\cdot \Bar{x} + \Bar{y}$ can be optimized by using a singular value decomposition (SVD), $\mathbf{USV}^\top = \mathcal{F}_\text{SVD}(\mathbf{H})$, where the  cross-covariance matrix $\mathbf{H}=\sum_{x_i\in \mathcal{X}_{in} }(x_i-\Bar{x})\cdot(y_{m(x_i)}-\Bar{y})^\top$,
$\Bar{x}=\frac{1}{|\mathcal{X}_{in}|}\sum_{x_i\in \mathcal{X}_{in}} x_i$ and $\Bar{y}=\frac{1}{|\mathcal{Y}_{in}|}\sum_{y_j\in \mathcal{Y}_{in}} y_j$.
To estimate the mapping $m$,
learning-based methods~\citep{wang2019deep,wang2019prnet,yew2020rpm,li2019iterative} match their encoded deep features:
\begin{equation} \label{eq:map}
    m(x_i,\: \mathcal{Y})=\mathcal{F}_{\text{map}}\big[\text{softmax}\big(\Theta_\mathcal{Y}\cdot{\Theta_{x_i}^{\top}}\big)\big],
\end{equation}
where $\mathcal{F_{\text{map}}[\cdot]}$ denotes a learnable mapping function, $\Theta_\mathcal{X} \in \mathbb{R}^{N\times C}$ and $\Theta_\mathcal{Y} \in \mathbb{R}^{M\times C}$ are embeded feature descriptors for each point in $\mathcal{X}$ and $\mathcal{Y}$, respectively.

\subsection{Pose-Invariant Features}
To estimate accurate correspondences between $\mathcal{X}$ and $\mathcal{Y}$, it is important to encode $\Theta_\mathcal{X}$ and $\Theta_\mathcal{Y}$ that are robust and even invariant to arbitrary transformations $\mathbf{T}
= [\mathbf{R},\: \mathbf{t}]
\in$ SE(3).
Previous networks~\citep{drost2010model,zhang-riconv-3dv19,chen2019clusternet} on 3D perception tasks can obtain SO(3)-invariance with the help of handcrafted {\textit{RI}} features, such as \textit{RRI}~\citep{chen2019clusternet} and \textit{PPF}~\citep{drost2010model}.

\noindent\textbf{Global {\textit{RI}} feature.}
As a global {\textit{RI}} feature, \textit{RRI} achieves pose-invariance~\footnote{Pose-invariance consists of rotation- and translation-invariance.} in accordance to a origin-centered unit sphere that usually uses the shape center of a point cloud as the coordinate center ``$O$'' (Fig.~\ref{fig:rot_feature} (a)).
Because complete point clouds under arbitrary transformations have the same shape center, its translation-invariance can be guaranteed by translating complete point clouds to the same origin-centered unit sphere coordinate.
However, different partial point clouds mostly have inconsistent shape centers caused by different missing parts, which challenges the pose-invariance of \textit{RRI}.

\noindent\textbf{Local {\textit{RI}} feature.}
Different from \textit{RRI}, 
\textit{PPF} features measure local pose-invariant point pairs relations, such as relative distances and surface normal deviations (Fig.~\ref{fig:rot_feature} (b)), which is invariant under arbitrary $\mathbf{R}\in$ SO(3) and $\mathbf{t}\in\mathbb{R}^3$.
Similar to \textit{PPF}, \textit{FPFH}~\citep{rusu2009fast} encodes pose-invariant geometric properties by using average curvature of the multi-dimensional histogram around a center point locally.
Both \textit{PPF} and \textit{FPFH} rely on the presence of consistent surface normal estimations.
Although we can estimate surface normals on-the-fly using local neighboring points, the surface normal direction ambiguities ({\color[RGB]{192,0,0}{red boxes}}) and 3D shape incompleteness ({\color[RGB]{255,192,0}{yellow boxes}}) make the normal estimation inconsistent, especially in the object-centric scenario (Fig.~\ref{fig:rot_feature} (c)).
Moreover, noisy measurements make 
pose-invariant correspondences estimation more challenging.

\begin{figure}
    \centering
    \includegraphics[width=0.9\linewidth]{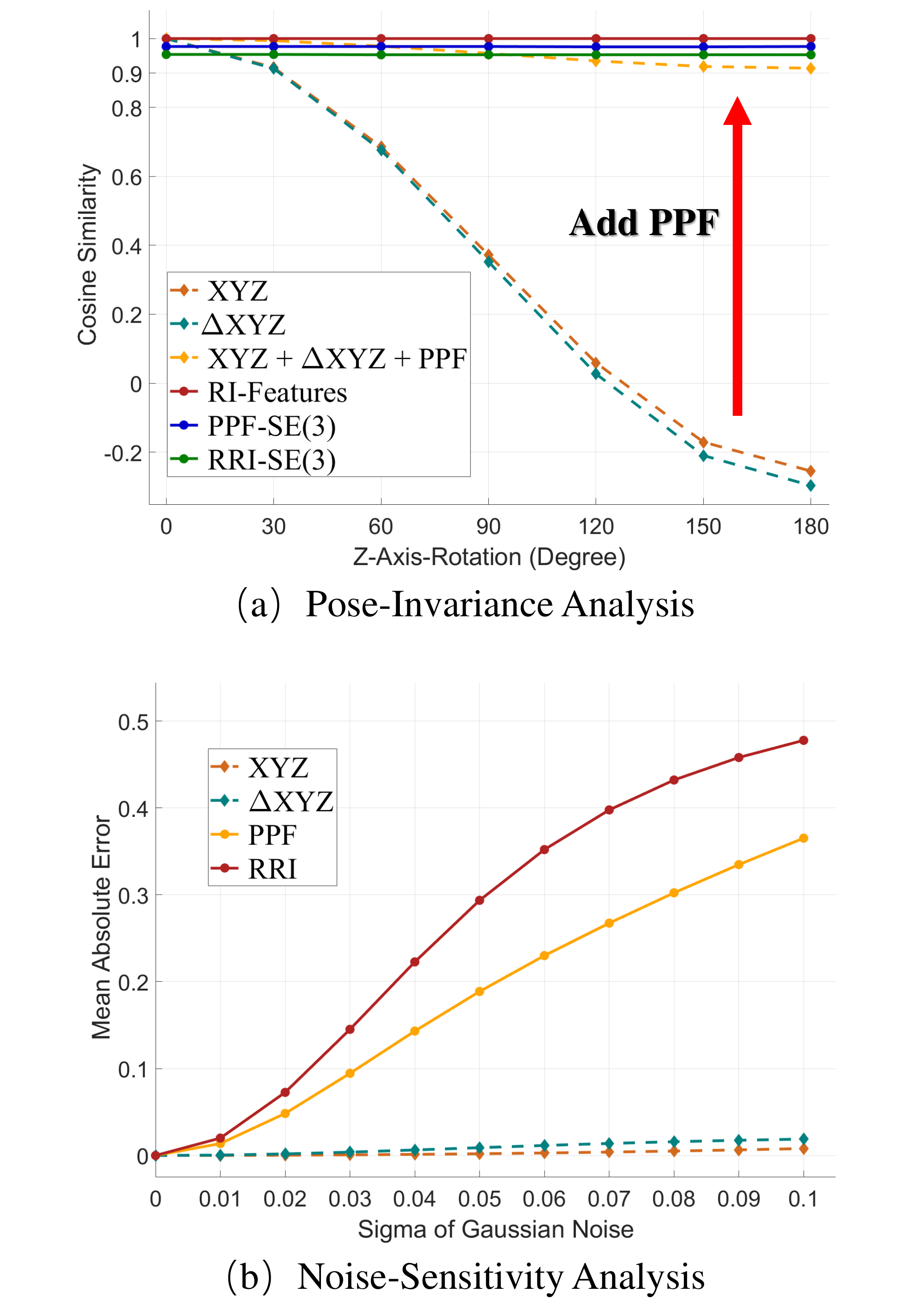}
    \caption{\textbf{Handcrafted Feature Analysis on Pose and Noise.} \textbf{(a)} 
    Handcrafted rotation-invariant features can highly increase rotation-robustness, but a SE(3) transformation (\emph{i.e.}, rotations and translations) can slightly influence their invariance.
    \textbf{(b)} Absolute position ``XYZ'' and relative position ``$\Delta$XYZ'' are more robust against noisy measurements than rotation-invariant features.
    }
    \label{fig:analysis}
\end{figure}
\subsection{Transformation-Robustness and Noise-Resilience}
To investigate robustness of various features against (a) arbitrary transformations and (b) random noise, 
we conduct a preliminary study (Fig.~\ref{fig:analysis}) on the ModelNet40 test set with 2,468 complete point clouds (2,048 points). 

\noindent\textbf{Rotation-Robustness.}
For rotation-robustness,
a rotation axis (\emph{e.g.}, Z-axis) is fixed, and then we gradually increase the rotation magnitude from 0\textdegree{} to 180\textdegree.
Handcrafted {\textit{RI}} features (\emph{e.g.}, \textit{RRI}, \textit{PPF} and \textit{FPFH}) are invariant against arbitrary rotations (see {\color[RGB]{178,34,34}{red line}}), but the absolute position ``XYZ'' and the relative position ``$\Delta$XYZ'' are highly influenced by the increment in the rotation magnitude.
Furthermore, concatenating the {\textit{RI}} features, such as \textit{PPF}, with ``XYZ'' and ``$\Delta$XYZ'' can highly increase 
the rotation-robustness~\footnote{Similar observations are reported in PPFNet (\emph{e.g.}, Table 6)~\citep{deng2018ppfnet}.} ({\color[RGB]{255,165,0}{yellow dash line}}).
However, if adding an arbitrary translation, the similarity scores ``PPF-SE(3)'' and ``RRI-SE(3)'' will be influenced (\emph{i.e.} \textless 1) due to ambiguous surface normal directions and mismatched shape centers, respectively.
\textit{FPFH} is similar with \textit{PPF}, and hence we 
omitted it in Fig.\ref{fig:analysis} (a).
Nonetheless, handcrafted {\textit{RI}} features can still describe {\textit{RI}} geometric properties (such as local curvatures and smoothness) for the corresponding observation, because
``PPF-SE(3)'' and ``RRI-SE(3)'' are invariant to arbitrary rotations, and showing two horizontal lines ({\color[RGB]{0,0,205}{blue line}} and {\color[RGB]{0,128,0}{green line}}) 
in Fig.~\ref{fig:analysis} (a).

\noindent\textbf{Noise-Resilience.}
As for noise-resilience test (Fig.~\ref{fig:analysis} (b)), we add a random Gaussian noise $\mathbb{\mathit{N}}(0,\, \sigma^2)$ to each point of point clouds, and gradually increases the $\sigma$ from 0 to 0.1.
The mean absolute error shows that handcrafted {\textit{RI}} features are much more influenced by noise than ``XYZ'' and ``$\Delta$XYZ'' by comparing each feature channel discrepancy with the increasing noise.
Hence, ``XYZ'' and ``$\Delta$XYZ'' could be more noise-resilient, which represents stable 3D geometrical structures for the point clouds.

\noindent\textbf{Analysis.}
According to our experiments, inconsistent measurements caused by center shifting, normal direction ambiguity or random noise, could influence the transformation-robustness of \textit{RI} features.
Although being robust against noisy measurements, global or relative coordinate features could be highly influenced by large transformations.
On the other side, coordinate-based features could serve as stable feature structure for organizing \textit{RI} features.
Moreover, we notice that many networks~\citep{zhang-riconv-3dv19, chen2019clusternet,li2021rotation,zhang2022riconv++} utilize \textit{RI} features along with random Gaussian noise for \textit{RI} classification, and it reveals that noise would not largely influence the shape semantics for grouped points.
Consequently, aggregating distribution-level features with {\textit{RI}} features and noise-resilient structural relations is promising to increase descriptor robustness in terms of arbitrary poses and random noise. 
However, previous methods~\citep{yuan2020deepgmr, deng2018ppfnet, yew2020rpm, li2019iterative} face difficulties in solving full-range PPR of noisy and incomplete point clouds. We argue that it is critical to construct network architecture that leverage the synergy between handcrafted {\textit{RI}} and coordinate features for transformation-robust and noise-resilient descriptors for global PPR.

\begin{figure*}
    \centering
    \includegraphics[width=1\linewidth]{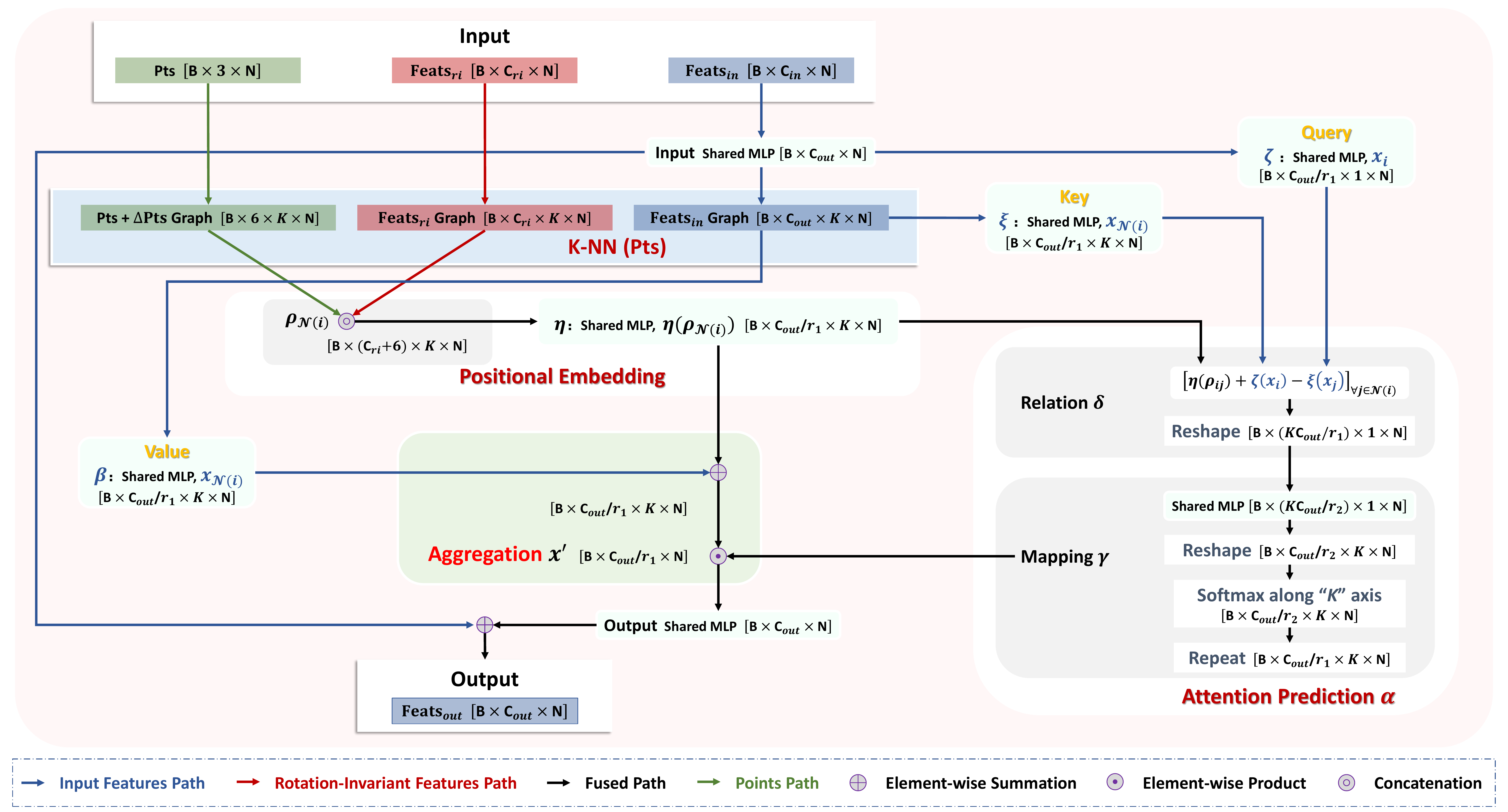}
    \caption{
    \textbf{Transformation-robust Point Transformer (TPT) module.}  TPT adaptively integrates point features by using various neighborhood graphs. 
    The feature tensor size is illustrated in detail, where ``B'', ``C'', ``\textit{K}'' and ``N'' are the number of batches, feature channels, neighboring points and total points, respectively.
    $r_1$ and $r_2$ are two hyper-parameters (\emph{e.g}, 4 and 8) for adjusting the 
    attention weights that can be shared across a group of channels.
    }
    \label{fig:PT}
\end{figure*}
\section{Our Approach}
We represent point cloud as a list of
neighborhood graphs $\{\mathcal{N}\}$, and each $\mathcal{N}=(\mathcal{V}, \mathcal{E}, \mathbf{X})$ consists of a node set $\mathcal{V}$ for spatial coordinates, an edge set $\mathcal{E}$ for neighboring connections and a node feature set $\mathbf{X}$ for encoded descriptors.
Consequently, the two partial point clouds $\mathcal{X}$ and $\mathcal{Y}$ are represented as two lists of neighborhood graphs $\{\mathcal{N}_s=(\mathcal{V}_s, \mathcal{E}_s, \mathbf{X})\}$ and $\{\mathcal{N}_t=(\mathcal{V}_t, \mathcal{E}_t, \mathbf{Y})\}$, 
and estimating pose-invariant correspondences between $\mathcal{X}$ and $\mathcal{Y}$ can be reformulated as a maximal common graph estimation problem between $\{\mathcal{N}_s\}$ and $\{\mathcal{N}_t\}$ by using noise-resilient graph structures and transformation-robust node features.

\begin{figure*}
    \centering
    \includegraphics[width=1\linewidth]{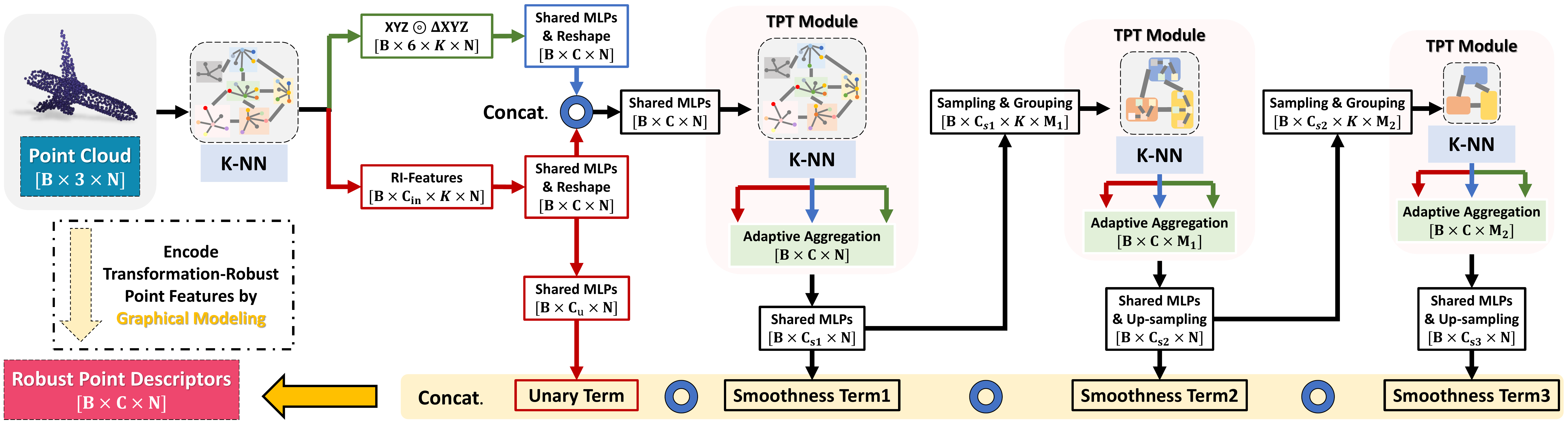}
    \caption{\textbf{Hierarchical Graphical Modeling for Encoding Transformation-Robust Feature Descriptors.}
    Following the graphical modeling structure, each point descriptor consists of 1) a ``Unary Term'' learned by the rotation-invariant feature graphs; 2) and multiple ``Smoothness Terms'' encoded from neighboring geometric features at different scales.
    }
    \label{fig:hgm}
\end{figure*}
\subsection{Transformation-Robust Point Transformer}
To encode robust point feature descriptors, we propose a novel Transformation-robust Point Transformer (\textbf{TPT}) module, which adaptively aggregates local features by considering both point feature similarities and their spatial positions for each neighborhood graph $\mathcal{N}$.
Notably, the neighborhood relationships $\mathcal{E}$ in point clouds are also pose-invariant.
In contrast to previous point cloud transformer operations~\citep{guo2021pct,zhao2021point} that are designed for 3D perception tasks, TPT emphasizes the pose-robustness for the aggregated point features by using handcrafted {\textit{RI}} features in the positional encoding.
The attention-based aggregation layer of TPT (Fig.~\ref{fig:PT}) can be formulated as follows:
\begin{equation}
    \mathbf{x}_i^{\prime} = \underset{j\in \mathcal{N}(i)}{\sum}
    \big(\beta(\mathbf{x}_j) + \eta(\rho_{ij})\big)
    \odot \alpha \big(\mathbf{x}_{\mathcal{N}(i)}, \eta(\rho_{\mathcal{N}(i)})\big)_j,
\end{equation}
where $\beta$ and $\eta$ are shared Multi-Layer-Perceptrons (MLPs), $\mathcal{N}(i)$ is the K-Nearest Neighboring (K-NN) point graph constructed in the spatial coordinates for the center point $x_i$, $\beta(\mathbf{x}_j)$ is the transformed features, $\eta(\rho_{ij})$ is the positional embedding, $\alpha \big(\mathbf{x}_{\mathcal{N}(i)}, \eta(\rho_{\mathcal{N}(i)})\big)_j$ denotes the attention weight $w_{ij} = \gamma\big(\delta(\mathbf{x}_{\mathcal{N}(i)}, \, \eta(\rho_{\mathcal{N}(i)})\big)_j$, and $\odot$ is element-wise product.
In view that vector attention operators~\citep{zhao2020exploring} often achieve better performance than scalar attention operators~\citep{vaswani2017attention}, we compute a vector attention that adapts to different feature channels, which consists of a relation operator $\delta$ and a mapping function $\gamma$:
\begin{equation}
    \delta(\mathbf{x}_{\mathcal{N}(i)},  \eta(\rho_{\mathcal{N}(i)})) = \big[[\zeta(\mathbf{x}_i) - \xi(\mathbf{x}_j) + \eta(\rho_{ij})]_{\forall j \in \mathcal{N}(i)} \big],
\end{equation}
where $\zeta$ and $\xi$ are shared MLPs, and $\delta$ combines all feature vectors by concatenation, which is equivalent to the ``Reshape'' operation in Fig.~\ref{fig:PT}.
Afterwards, $\gamma$ further transforms the set of vectors, and then maps to the right dimensionality.
Particularly, we carefully design the positional embedding for describing the positional relations between pairwise connected points by concatenating the absolute positions, relative positions and handcrafted {\textit{RI}} features (\emph{e.g.} \textit{RRI} and \textit{PPF}).
Although adding absolute and relative positions may influence the pose-invariance, it further increases PPR performance against noise.

\subsection{Hierarchical Graphical Modeling}
A probabilistic graphical model utilizes a graph-based representation to express conditional dependency among random variables, which can encode a distribution over a multi-dimensional space with the graph structure~\citep{koller2009probabilistic}.
Given that
a point cloud can be represented as a set of neighborhood graphs $\{\mathcal{N}\} = \{(\mathcal{V}, \mathcal{E}, \mathbf{X})\}$, 
we propose a Hierarchical Graphical Modeling (HGM) architecture to encode multi-scale distribution features for constructing robust point feature descriptors for the source and the target individually.
As shown in Fig.~\ref{fig:hgm}, the constructed descriptors for each point consist of 1) a unary term focusing on each node learned by handcrafted {\textit{RI}} features graphs;
2) multiple smoothness terms focusing on 3D geometric distributions
encoded from multi-scale structural relations using the proposed TPT modules (similar with the \textbf{\textit{Sum-Product}}~\citep{koller2009probabilistic}).
Consequently, the encoded descriptors for a point point $x_i$ can be formulated as:
\begin{equation}\begin{split}
    \Theta_{x_i} = 
    \Big[
    \Omega\big(\mathbf{x}_{\mathcal{N}_1{(i)}}^{\text{RI}}\big),
    \Lambda^1\big(\mathbf{x}_{\mathcal{N}_1{(i)}}\big),
    \Lambda^2\big(\mathbf{x}_{\mathcal{N}_2{(i)}}\big),
    \Lambda^3\big(\mathbf{x}_{\mathcal{N}_3{(i)}}\big)
    \Big],
\end{split}\end{equation}
where $\Theta_{x_i}$ concatenates the unary term 
$\Omega\big(\cdot)$
encoded by {\textit{RI}} features $\mathbf{x}_{\mathcal{N}_1{(i)}}^{\text{RI}}$, 
and the smoothness terms 
$\Lambda^1(\cdot)$, $\Lambda^2(\cdot)$ and $\Lambda^3(\cdot)$.
$\mathcal{N}_1{(i)}$, $\mathcal{N}_2{(i)}$ and $\mathcal{N}_3{(i)}$ 
are neighborhood graphs at different scales.
Accordingly, the mapping function $m$ estimation can be formulated as an assignment problem with the following objective function:
\begin{equation}
    \begin{split}
        E(m) = \sum_{x_i\in \mathcal{X}}\Big(&\lambda_{u}E_u\big(\Omega(\mathbf{x}_{\mathcal{N}_1{(i)}}^{\text{RI}}), \Omega(\mathbf{Y}_{\mathcal{N}_1}^{\text{RI}})\big) \\
        + \sum_{n}&\lambda_{s_n}E_{s_n}\big(\Lambda^n(\mathbf{x}_{\mathcal{N}_n{(i)}}), \Lambda^n(\mathbf{Y}_{\mathcal{N}_n}) \big)\Big),
    \end{split}
\end{equation}
where $E_u(\cdot)$ denotes the unary penalty term, $E_{s_n}(\cdot)$ denotes each smoothness penalty term, and $\lambda(\cdot)$ is the corresponding weights for each term.

{Unlike previous methods~\citep{deng2018ppfnet,yew2020rpm} that only use the nearest neighbor as the paired points to encode handcrafted {\textit{RI}} features,
we consider each point as a centroid that is paired with more neighboring points for $\Omega\big(\cdot)$.
We highlight that our unary term $\Omega\big(\cdot)$ rigorously preserves rotation-invariance, which can resolve full-range PCR without smoothness terms if the measurements are clean and consistent.
The smoothness terms $\Lambda^n(\cdot)$ is transformation-robust with the help of {\textit{RI}} features, which highly increases noise-resilience for optimizing the mapping function $m$.
}

\begin{figure*}
    \centering
    \includegraphics[width=0.7\linewidth]{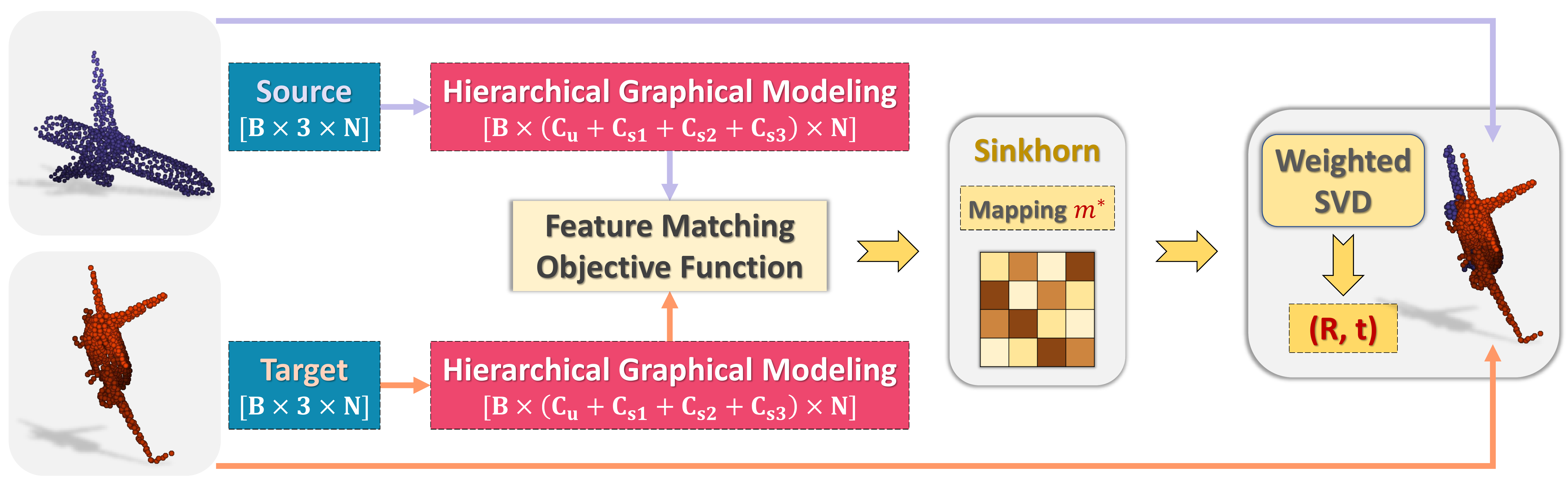}
    \caption{
    \textbf{Framework Overview.}
    \OM{} estimates pose-invariant correspondences using the robust feature descriptors encoded by our HGM.
    Note that the two HGM share weights.
    }
    \label{fig:gmc}
\end{figure*}
\subsection{Optimization}
\noindent\textbf{Mapping Function Optimization.}
We penalize the mis-matching loss by encouraging correspondences with similar feature descriptors. 
For each potential mapping $m(x_i, y_j)$, we compute element-wise feature distances between corresponding terms from $\Theta_{\mathcal{X}}$ and $\Theta_{\mathcal{Y}}$:
\begin{equation}
    \begin{split}
        E&(m_{ij}) = \sum_\mathbf{C}\Big( \lambda_{u}\big(\big\|\Omega(\mathbf{x}_{\mathcal{N}_1{(i)}}^{\text{RI}}) - \Omega(\mathbf{y}_{\mathcal{N}_1{(j)}}^{\text{RI}})\big\|_2^2\big /\sqrt{|\mathbf{C}_u|}\big) \\ 
        &+ \sum_{n}\lambda_{s_n}\big(\big\|\Lambda^n(\mathbf{x}_{\mathcal{N}_n{(i)}}) - \Lambda^n(\mathbf{y}_{\mathcal{N}_n{(j)}})\big\|_2^2 \big /\sqrt{|\mathbf{C}_{s_n}|} \big)\Big), 
    \end{split}
\end{equation}
where $|\mathbf{C_u}|$ and $|\mathbf{C_{s_n}}|$ are the feature size for each corresponding feature item, which adjusts their importance.
Hence, $m(x_i, y_j)$ is initialized as $ \exp\big({{-E(m_{ij})}}\big)$.
Thereafter, we use Sinkhorn normalization layers for outlier rejection~\citep{santa2017deeppermnet, sarlin2020superglue, yew2020rpm}.
With the optimized matching matrix $m^{\star}$, the  cross-covariance matrix $\mathbf{H}=\sum_{i=1}^{N} c_i\cdot(x_i-\Bar{x})\cdot(y[{m^{\star}(x_i)}]-\Bar{y})^\top$,
where $\Bar{x}=\sum_{i=1}^{N} c_i \cdot x_i$, $\Bar{y}=\sum_{i=1}^{N} c_i \cdot y[{m^{\star}(x_i)}]$,
$y[{m^{\star}(x_i)}] =\frac{1}{\sum_jm^{\star}(x_i, y_j)} \sum_j^M y_j\cdot m^{\star}(x_i, y_j)$, 
and $c_i = \sum_j^M m^{\star}(x_i, y_j)$,
which denotes the confidence for point $x_i$ is an inlier.
In the end, $\mathbf{R}_{\mathcal{XY}}$
and $\mathbf{t}_{\mathcal{XY}}$ can be resolved by a SVD introduced in Sec.~\ref{sec:pa}.
Consequently, our \OM{} framework overview can be illustrated in Fig.~\ref{fig:gmc}.

\noindent\textbf{Training Loss.}
\OM{} is trained end-to-end. 
Following~\citep{yew2020rpm}, the training loss consists of two parts,
a registration loss $\mathcal{L}_{\mathrm{RG}}$
and an auxiliary loss $\mathcal{L}_{\mathrm{IL}}$ to encourage inliers.
In addition, we further improves registration estimation by using a cycle loss function that is formulated as:
\begin{equation}
    \mathcal{L} = \mathcal{L}_{\mathrm {RG}} + \omega\mathcal{L}_{\mathrm{IL}} = \mathcal{L}_{\mathrm{reg}}^{\mathcal{X}\mathcal{Y}} + \mathcal{L}_{\mathrm{reg}}^{\mathcal{Y}\mathcal{X}} + \omega(\mathcal{L}_{\mathrm{inlier}}^{\mathcal{X}\mathcal{Y}}+\mathcal{L}_{\mathrm{inlier}}^{\mathcal{Y}\mathcal{X}}),
\end{equation}
where $\mathcal{L}_{\mathrm{reg}}^{\mathcal{X} \mathcal{Y}} = \frac{1}{N}\sum_{i}^{N}|\mathbf{{T}}^{\mathcal{XY}}_{\text{pred.}}(x_i) - \mathbf{{T}}^{\mathcal{XY}}_{\text{GT}}(x_i)|$, 
$\mathbf{{T}}^{\mathcal{XY}}_{\text{pred.}}$ is our predicted transformation for $\mathcal{X}\xrightarrow{}\mathcal{Y}$,
$\mathcal{L}_{\mathrm{inlier}}^{\mathcal{X}\mathcal{Y}}=\frac{1}{M}\sum_j^M(1-\sum_i^N m^{\star}(x_i,y_j)) + \frac{1}{N}\sum_i^N(1-\sum_j^M m^{\star}(x_i,y_j))$. 
$\mathcal{L}_{\mathrm{reg}}^{\mathcal{Y}\mathcal{X}}$ and 
$\mathcal{L}_{\mathrm{inlier}}^{\mathcal{Y}\mathcal{X}}$ are defined with the opposite direction $\mathcal{Y}\xrightarrow{}\mathcal{X}$ for $\mathcal{L}_{\mathrm{reg}}^{\mathcal{X} \mathcal{Y}}$ and $\mathcal{L}_{\mathrm{inlier}}^{\mathcal{X}\mathcal{Y}}$, respectively.

\begin{figure}
    \centering
    \includegraphics[width=1\linewidth]{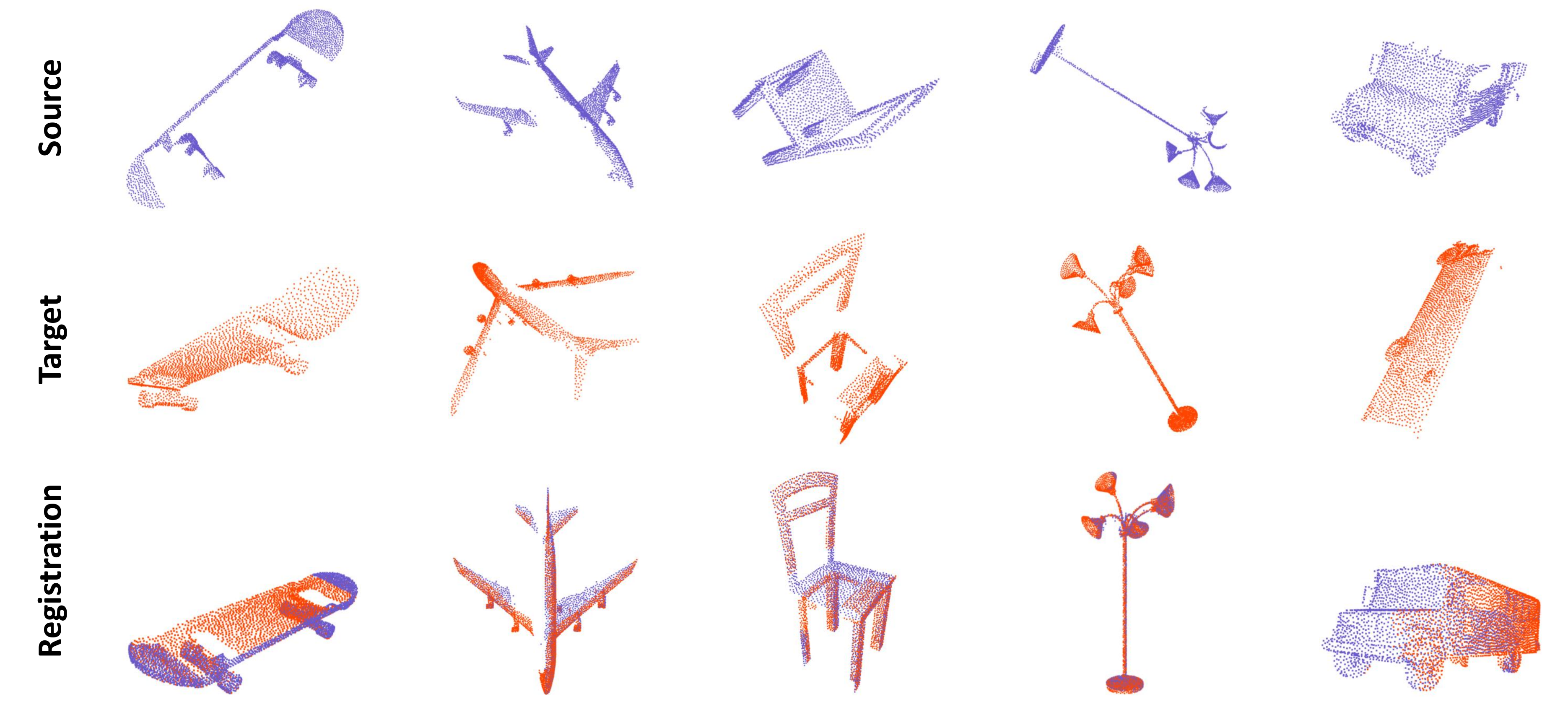}
    \caption{\textbf{Data examples in MVP-RG dataset.} The source point clouds and the target are different partial scans under a relative transformation in SE(3) for the same object.}
    \label{fig:mvp}
\end{figure}
\begin{figure*}
    \centering
    \vspace{-2mm}
    \includegraphics[width=1\linewidth]{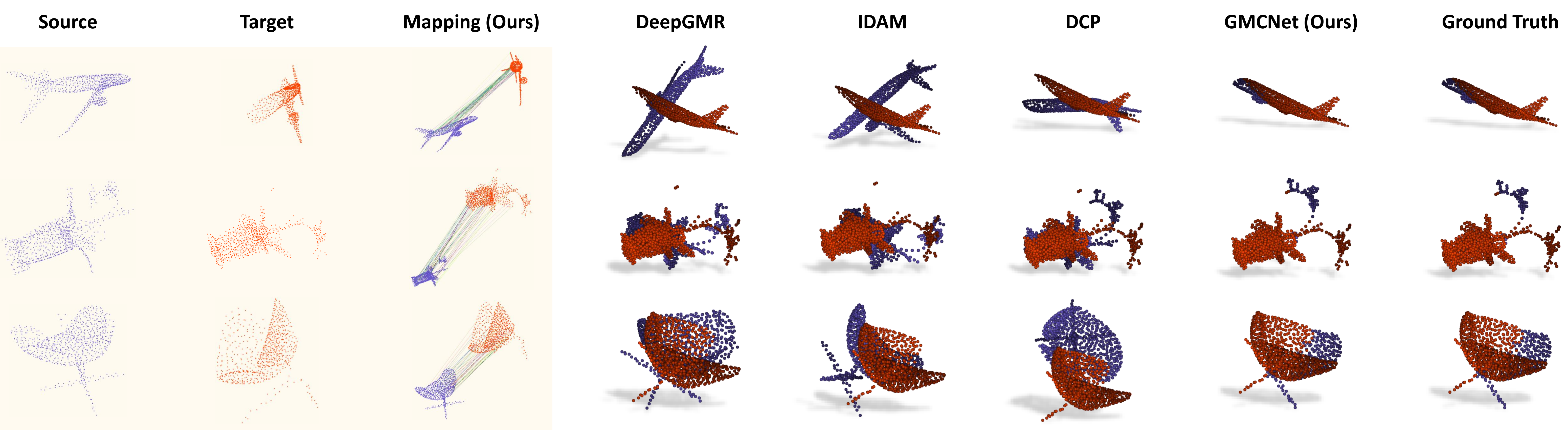}
    \caption{
    \textbf{Qualitative Partial-to-Partial Point Cloud Registration Results on ModelNet40}~\citep{wu20153d}.
    50 randomly selected correspondences by \OM{} are visualized in ``Mapping''.
    }
    \label{fig:mn40}
\end{figure*}
\section{Virtual Scanned MVP-RG Dataset}
\noindent\textbf{Existing Benchmark.}
First of all, we highlight that object-centric PCR is different from large-scale raw-scanned point cloud registration for the following main factors:
\textbf{1)} They are mainly created for indoor or outdoor scene-level 3D point clouds, and hence are much larger-scale than object-centric PCR.
\textbf{2)} They often have much more points (\emph{e.g.} often over 30,000 points), but object-centric PPR often has around 2,000 points.
\textbf{3)} They mostly have small transformations between consecutive frames.
Most existing methods on object-centric PCR are evaluated on ModelNet40~\citep{wu20153d}, which consists of complete point clouds uniformly sampled from CAD model surfaces.
However, their uniformly distributed points are far different from real observations that are biased by current camera pose.
Specifically, view-based scans will mostly lead to partial observations and inconsistent local point densities, which makes the registration more challenging.
Recently, the MVP dataset~\citep{pan2021variational} consisting of multi-view virtual-scanned partial point clouds is proposed for point cloud completion, and it generates diverse partial observations for each 3D CAD model from 26 uniformly-distributed camera poses on a unit sphere.
Because there are no large-scale real-scanned point clouds for object-centric registration, the proposed MVP-RG dataset makes an important step forward, which could facilitate future research on solving the sparse, noisy and challenging object-centric PPR problem.

\noindent\textbf{Constructing MVP-RG.}
Following the MVP dataset, we establish a challenging Multi-View Partial virtual scan ReGistration (MVP-RG) dataset (see Fig.~\ref{fig:teaser} (a)) for PPR.
In the MVP-RG dataset, each partial point cloud is generated by projecting the scanned depth map to the image coordinate frame and followed by a FPS to 2,048 points.
Afterwards, we select paired of partial point clouds for the same object if sufficient overlapping areas are detected. 
In total, our MVP-RG dataset consists of 7,600 partial point cloud pairs from 16 categories,
which is split into a training set (6,400 samples) and a testing set (1,200 samples).

\noindent\textbf{Properties and Advantages.}
Different from the MVP dataset that assumes all objects are in the same canonical pose, we randomly transform the 3D object for each observation,
and hence making it a full-range PPR problem.
Partial point cloud pair examples are shown in Fig.~\ref{fig:mvp}.
Particularly, the MVP-RG settings can better adapt to 6D-pose object estimation tasks, as they often utilize a multi-camera system (\emph{e.g.} BigBIRD~\citep{singh2014bigbird}) for estimating object pose in the full SE(3) domain.

\section{Experiments}
\noindent\textbf{Datasets.}
We evaluate \OM{} on partial point clouds from the ModelNet40~\citep{wu20153d} and MVP-RG datasets.
We evaluate both \textit{restricted} rotations (\emph{i.e.} [0, 45\textdegree]) and unrestricted rotations (\emph{i.e.} SO(3)) defined by an arbitrary rotation axis~\footnote{Opposite axis directions are equivalent to rotations in [-180\textdegree, 0]} and a random rotation magnitude in [0, 180\textdegree]~\citep{yuan2020deepgmr}.
For all evaluated benchmarks, \OM{} achieves superior performance over previous learning-based methods.

\noindent\textbf{Implementation Details.}
Our networks are implemented using PyTorch.
We use Open3D library~\citep{zhou2018open3d} to implement normal estimation on-the-fly.
And we set all surface normal directions orient towards the coordinate origin of the posed partial point cloud.
Our models are trained using the Adam optimizer~\citep{kingma2014adam} with an initial learning rate 
0.001
on an NVIDIA TITAN Xp GPU for most experiments.

\begin{table*}[]
    \setlength{\tabcolsep}{5.5pt}
    \caption{
    \textbf{Partial-to-Partial Point Cloud Registration on ModelNet40 (Clean and Unseen).} 
    $^{\dag}$ denotes ground truth correspondences or cross-contextual information are used during training.
    }
    \begin{center}
        \begin{tabular}{l|ccc|ccc|ccc|ccc}
            \toprule[1.2pt]
            \multicolumn{1}{c}{} & \multicolumn{6}{c|}{\fontsize{8}{8}\selectfont{Clean}} &
            \multicolumn{6}{c}{\fontsize{8}{8}\selectfont{Unseen}} \\
            \cmidrule[0.25pt]{2-13}
            \multicolumn{1}{c}{} & \multicolumn{3}{c|}{\fontsize{8}{8}\selectfont{[0, 45\text{\textdegree}]}} & \multicolumn{3}{c|}{\fontsize{8}{8}\selectfont{[0, 180\text{\textdegree}]}} & \multicolumn{3}{c|}{\fontsize{8}{8}\selectfont{[0, 45\text{\textdegree}]}} & \multicolumn{3}{c}{\fontsize{8}{8}\selectfont{[0, 180\text{\textdegree}]}} \\
            \cmidrule[0.25pt]{2-13}
            \multicolumn{1}{c}{} & \scriptsize $\mathcal{L}_{\mathbf{R}}$   & \scriptsize $\mathcal{L}_{\mathbf{t}}$   & \scriptsize $\mathcal{L}_{\text{RMSE}}$   & \scriptsize $\mathcal{L}_{\mathbf{R}}$   & \scriptsize $\mathcal{L}_{\mathbf{t}}$   & \scriptsize $\mathcal{L}_{\text{RMSE}}$ & \scriptsize $\mathcal{L}_{\mathbf{R}}$   & \scriptsize $\mathcal{L}_{\mathbf{t}}$   & \scriptsize $\mathcal{L}_{\text{RMSE}}$   & \scriptsize $\mathcal{L}_{\mathbf{R}}$   & \scriptsize $\mathcal{L}_{\mathbf{t}}$   & \scriptsize $\mathcal{L}_{\text{RMSE}}$ \\
            \midrule[0.75pt] 
            \scriptsize PRNet$^{\dag}$~\citep{wang2019prnet} & \fontsize{6.5}{6.5}\selectfont 8.00\textdegree & \fontsize{6.5}{6.5}\selectfont 0.054 & \fontsize{6.5}{6.5}\selectfont 0.073 &  \fontsize{6.5}{6.5}\selectfont 98.16\textdegree & \fontsize{6.5}{6.5}\selectfont 0.325 & \fontsize{6.5}{6.5}\selectfont 0.543 & \fontsize{6.5}{6.5}\selectfont 3.19\textdegree & \fontsize{6.5}{6.5}\selectfont 0.028 & \fontsize{6.5}{6.5}\selectfont 0.036 &  \fontsize{6.5}{6.5}\selectfont 91.94\textdegree & \fontsize{6.5}{6.5}\selectfont 0.297 & \fontsize{6.5}{6.5}\selectfont 0.545 \\
            \scriptsize IDAM$^{\dag}$({GNN})~\citep{li2019iterative}& \fontsize{6.5}{6.5}\selectfont 1.66\textdegree & \fontsize{6.5}{6.5}\selectfont 0.009 & \fontsize{6.5}{6.5}\selectfont 0.013 & \fontsize{6.5}{6.5}\selectfont 25.36\textdegree & \fontsize{6.5}{6.5}\selectfont 0.126 & \fontsize{6.5}{6.5}\selectfont 0.166 & \fontsize{6.5}{6.5}\selectfont 1.61\textdegree & \fontsize{6.5}{6.5}\selectfont 0.010 & \fontsize{6.5}{6.5}\selectfont 0.014 & \fontsize{6.5}{6.5}\selectfont 20.80\textdegree & \fontsize{6.5}{6.5}\selectfont 0.112 & \fontsize{6.5}{6.5}\selectfont 0.146 \\
            \scriptsize IDAM$^{\dag}$(\textit{FPFH})~\citep{li2019iterative} & \fontsize{6.5}{6.5}\selectfont 0.93\textdegree & \fontsize{6.5}{6.5}\selectfont 0.005 & \fontsize{6.5}{6.5}\selectfont 0.007 & \fontsize{6.5}{6.5}\selectfont 18.06\textdegree & \fontsize{6.5}{6.5}\selectfont 0.083 & \fontsize{6.5}{6.5}\selectfont 0.115 & \fontsize{6.5}{6.5}\selectfont 0.86\textdegree & \fontsize{6.5}{6.5}\selectfont 0.005 & \fontsize{6.5}{6.5}\selectfont 0.007 & \fontsize{6.5}{6.5}\selectfont 16.17\textdegree & \fontsize{6.5}{6.5}\selectfont 0.073 & \fontsize{6.5}{6.5}\selectfont 0.106  \\
            \scriptsize RGM$^{\dag}$~\citep{fu2021robust} & \fontsize{6.5}{6.5}\selectfont 0.30\textdegree & \fontsize{6.5}{6.5}\selectfont 0.002 & \fontsize{6.5}{6.5}\selectfont 0.003 & \fontsize{6.5}{6.5}\selectfont 24.10\textdegree & \fontsize{6.5}{6.5}\selectfont 0.115 & \fontsize{6.5}{6.5}\selectfont 0.160 & \fontsize{6.5}{6.5}\selectfont 0.34\textdegree & \fontsize{6.5}{6.5}\selectfont 0.004 & \fontsize{6.5}{6.5}\selectfont 0.004 & \fontsize{6.5}{6.5}\selectfont 8.78\textdegree & \fontsize{6.5}{6.5}\selectfont 0.076 & \fontsize{6.5}{6.5}\selectfont 0.084  \\
            \scriptsize Predator$^{\dag}$~\citep{huang2021predator} & \fontsize{6.5}{6.5}\selectfont 1.60\textdegree & \fontsize{6.5}{6.5}\selectfont 0.012 & \fontsize{6.5}{6.5}\selectfont 0.015 & \fontsize{6.5}{6.5}\selectfont 15.33\textdegree & \fontsize{6.5}{6.5}\selectfont 0.041 & \fontsize{6.5}{6.5}\selectfont 0.077 & \fontsize{6.5}{6.5}\selectfont 1.32\textdegree & \fontsize{6.5}{6.5}\selectfont 0.009 & \fontsize{6.5}{6.5}\selectfont 0.012 & \fontsize{6.5}{6.5}\selectfont 11.59\textdegree & \fontsize{6.5}{6.5}\selectfont 0.032 & \fontsize{6.5}{6.5}\selectfont 0.058  \\
            \scriptsize DCP~\citep{wang2019deep} & \fontsize{6.5}{6.5}\selectfont 10.24\textdegree & \fontsize{6.5}{6.5}\selectfont 0.071 & \fontsize{6.5}{6.5}\selectfont 0.103 &  \fontsize{6.5}{6.5}\selectfont 70.51\textdegree & \fontsize{6.5}{6.5}\selectfont 0.186 & \fontsize{6.5}{6.5}\selectfont 0.425 & \fontsize{6.5}{6.5}\selectfont 11.92\textdegree & \fontsize{6.5}{6.5}\selectfont 0.076 & \fontsize{6.5}{6.5}\selectfont 0.119 &  \fontsize{6.5}{6.5}\selectfont 67.39\textdegree & \fontsize{6.5}{6.5}\selectfont 0.170 & \fontsize{6.5}{6.5}\selectfont 0.410  \\
            \scriptsize DeepGMR (\textit{RRI})~\citep{yuan2020deepgmr} & \fontsize{6.5}{6.5}\selectfont 16.95\textdegree & \fontsize{6.5}{6.5}\selectfont 0.070 & \fontsize{6.5}{6.5}\selectfont 0.122 &  \fontsize{6.5}{6.5}\selectfont 45.38\textdegree & \fontsize{6.5}{6.5}\selectfont 0.210 & \fontsize{6.5}{6.5}\selectfont 0.321 & \fontsize{6.5}{6.5}\selectfont 17.45\textdegree & \fontsize{6.5}{6.5}\selectfont 0.074 & \fontsize{6.5}{6.5}\selectfont 0.130 & \fontsize{6.5}{6.5}\selectfont 49.23\textdegree & \fontsize{6.5}{6.5}\selectfont 0.219 & \fontsize{6.5}{6.5}\selectfont 0.349  \\
            \scriptsize DeepGMR (XYZ)~\citep{yuan2020deepgmr} & \fontsize{6.5}{6.5}\selectfont 6.60\textdegree & \fontsize{6.5}{6.5}\selectfont 0.050 & \fontsize{6.5}{6.5}\selectfont 0.065 & \fontsize{6.5}{6.5}\selectfont 67.81\textdegree & \fontsize{6.5}{6.5}\selectfont 0.244 & \fontsize{6.5}{6.5}\selectfont 0.416 & \fontsize{6.5}{6.5}\selectfont 8.05\textdegree & \fontsize{6.5}{6.5}\selectfont 0.053 & \fontsize{6.5}{6.5}\selectfont 0.074 & \fontsize{6.5}{6.5}\selectfont 72.27\textdegree & \fontsize{6.5}{6.5}\selectfont 0.245 & \fontsize{6.5}{6.5}\selectfont 0.449  \\
            \scriptsize RPMNet (\textit{PPF})~\citep{yew2020rpm} & \fontsize{6.5}{6.5}\selectfont 0.78\textdegree & \fontsize{6.5}{6.5}\selectfont 0.005 & \fontsize{6.5}{6.5}\selectfont 0.006 & \fontsize{6.5}{6.5}\selectfont 15.84\textdegree & \fontsize{6.5}{6.5}\selectfont 0.070 & \fontsize{6.5}{6.5}\selectfont 0.098 & \fontsize{6.5}{6.5}\selectfont 0.60\textdegree & \fontsize{6.5}{6.5}\selectfont 0.004 & \fontsize{6.5}{6.5}\selectfont 0.005 & \fontsize{6.5}{6.5}\selectfont 16.91\textdegree & \fontsize{6.5}{6.5}\selectfont 0.079 & \fontsize{6.5}{6.5}\selectfont 0.127  \\
            \midrule[0.25pt]
            \scriptsize \OM{} (\textit{RRI})  & \fontsize{6.5}{6.5}\selectfont 0.013\textdegree & \fontsize{6.5}{6.5}\selectfont \textless 0.0001 & \fontsize{6.5}{6.5}\selectfont \textless 0.0001 & \fontsize{6.5}{6.5}\selectfont \textbf{0.036\textdegree} & \fontsize{6.5}{6.5}\selectfont \textbf{0.0002} & \fontsize{6.5}{6.5}\selectfont \textbf{0.0002} & \fontsize{6.5}{6.5}\selectfont 0.017\textdegree & \fontsize{6.5}{6.5}\selectfont \textless 0.0001 & \fontsize{6.5}{6.5}\selectfont \textless 0.0001 & \fontsize{6.5}{6.5}\selectfont \textbf{0.053\textdegree} & \fontsize{6.5}{6.5}\selectfont \textbf{0.0004} & \fontsize{6.5}{6.5}\selectfont \textbf{0.0005}  \\
            \scriptsize \OM{} (\textit{FPFH})  & \fontsize{6.5}{6.5}\selectfont \textbf{0.012\textdegree} & \fontsize{6.5}{6.5}\selectfont \textbf{\textless 0.0001} & \fontsize{6.5}{6.5}\selectfont \textbf{\textless 0.0001} & \fontsize{6.5}{6.5}\selectfont 0.253\textdegree & \fontsize{6.5}{6.5}\selectfont 0.0008 & \fontsize{6.5}{6.5}\selectfont 0.0016 & \fontsize{6.5}{6.5}\selectfont \textbf{0.014\textdegree} & \fontsize{6.5}{6.5}\selectfont \textbf{\textless 0.0001} & \fontsize{6.5}{6.5}\selectfont \textbf{\textless 0.0001} & \fontsize{6.5}{6.5}\selectfont 0.549\textdegree & \fontsize{6.5}{6.5}\selectfont 0.0024 & \fontsize{6.5}{6.5}\selectfont 0.0033  \\
            \scriptsize \OM{} (\textit{PPF})  & \fontsize{6.5}{6.5}\selectfont 0.016\textdegree & \fontsize{6.5}{6.5}\selectfont \textless 0.0001 & \fontsize{6.5}{6.5}\selectfont \textless 0.0001 & \fontsize{6.5}{6.5}\selectfont 0.284\textdegree & \fontsize{6.5}{6.5}\selectfont 0.0011 & \fontsize{6.5}{6.5}\selectfont 0.0017 & \fontsize{6.5}{6.5}\selectfont 0.026\textdegree & \fontsize{6.5}{6.5}\selectfont 0.0002 & \fontsize{6.5}{6.5}\selectfont 0.0002 & \fontsize{6.5}{6.5}\selectfont 0.388\textdegree & \fontsize{6.5}{6.5}\selectfont 0.0024 & \fontsize{6.5}{6.5}\selectfont 0.0030  \\
            \bottomrule[1.2pt]
        \end{tabular}
    \end{center}
    \label{tab:mn40_clean_and_unseen}
\end{table*}

\noindent\textbf{Evaluation Metrics.}
Following~\citep{yew2020rpm, yuan2020deepgmr}, we use mean isotropic rotation errors 
$\mathcal{L}_\mathbf{R}$, 
translation errors 
$\mathcal{L}_\mathbf{t}$,
and the root mean square error 
$\mathcal{L}_{\text{RMSE}}$ 
that evaluates registration in the context of scene reconstruction:
$\mathcal{L}_\mathbf{R} = \arccos\big(\frac{1}{2}(tr(\mathbf{R}_{\text{GT}}^{-1} \cdot \mathbf{R}_{\text{pred.}})-1)\big)$,
$\mathcal{L}_\mathbf{t} = \big\| \mathbf{t}_{\text{GT}} - \mathbf{{t}}_{\text{pred.}} \big\|_2$,
and 
$\mathcal{L}_{\text{RMSE}} = \frac{1}{N}\sqrt{\sum_{i=1}^N\Big\|\mathbf{T}_{\text{GT}}(x_i) - \mathbf{{T}}_{\text{pred.}}(x_i)\Big\|^2}$.
Note that we use all points to compute $\mathcal{L}_{\text{RMSE}}$.

\noindent\textbf{\textit{RI} Features.}
\textbf{1) \textit{RRI}} features require at least two neighbors~\citep{chen2019clusternet} (you can check their original paper, and in their Eq. (6), $j\neq k$), and therefore we use two neighbors to encode \textit{RRI} features (feature channel size $C = 8$) in the \textit{RI} feature graphs of TPT modules.
\textbf{2) \textit{FPFH}} features are constructed by using many neighboring points, and its feature channel size $C = 33$.
We use the implementation by open3D~\citep{zhou2018open3d}.
The \textit{FPFH} features are only used for the input node features, and we use \textit{PPF} for TPT modules, because it is very inefficient to construct FPFH for all network layers.
\textbf{3) \textit{PPF}} uses the estimated surface normals on-the-fly by open3D~\citep{zhou2018open3d}. 
For the \textit{RI} feature graphs in TPT modules, each point pair is used to encode one \textit{PPF} ($C = 4$).

\begin{table*}[]
    \setlength{\tabcolsep}{7pt}
    \caption{
    \textbf{Partial-to-Partial Point Cloud Registration on ModelNet40 (Noisy and Unseen).} 
    $^{\dag}$ denotes ground truth correspondences or cross-contextual information are used during training.
    $NC$ denotes Not-Converge, and it means the network does not converge during training.
    }
    \begin{center}
        \begin{tabular}{l|ccc|ccc|ccc|ccc}
            \toprule[1.2pt]
            \multicolumn{1}{c}{} & \multicolumn{6}{c|}{\fontsize{8}{8}\selectfont{Noisy}} &
            \multicolumn{6}{c}{\fontsize{8}{8}\selectfont{Noisy $\&$ Unseen}} \\
            \cmidrule[0.25pt]{2-13}
            \multicolumn{1}{c}{} & \multicolumn{3}{c|}{\fontsize{8}{8}\selectfont{[0, 45\text{\textdegree}]}} & \multicolumn{3}{c|}{\fontsize{8}{8}\selectfont{[0, 180\text{\textdegree}]}} & \multicolumn{3}{c|}{\fontsize{8}{8}\selectfont{[0, 45\text{\textdegree}]}} & \multicolumn{3}{c}{\fontsize{8}{8}\selectfont{[0, 180\text{\textdegree}]}} \\
            \cmidrule[0.25pt]{2-13}
            \multicolumn{1}{c}{} & \scriptsize $\mathcal{L}_{\mathbf{R}}$   & \scriptsize $\mathcal{L}_{\mathbf{t}}$   & \scriptsize $\mathcal{L}_{\text{RMSE}}$   & \scriptsize $\mathcal{L}_{\mathbf{R}}$   & \scriptsize $\mathcal{L}_{\mathbf{t}}$   & \scriptsize $\mathcal{L}_{\text{RMSE}}$ & \scriptsize $\mathcal{L}_{\mathbf{R}}$   & \scriptsize $\mathcal{L}_{\mathbf{t}}$   & \scriptsize $\mathcal{L}_{\text{RMSE}}$   & \scriptsize $\mathcal{L}_{\mathbf{R}}$   & \scriptsize $\mathcal{L}_{\mathbf{t}}$   & \scriptsize $\mathcal{L}_{\text{RMSE}}$ \\
            \midrule[0.75pt] 
            \scriptsize PRNet$^{\dag}$~\citep{wang2019prnet} & \fontsize{6.5}{6.5}\selectfont 4.37\textdegree & \fontsize{6.5}{6.5}\selectfont 0.034 & \fontsize{6.5}{6.5}\selectfont 0.045 & \fontsize{6.5}{6.5}\selectfont 95.80\textdegree & \fontsize{6.5}{6.5}\selectfont 0.319 & \fontsize{6.5}{6.5}\selectfont 0.542 &
            \fontsize{6.5}{6.5}\selectfont 8.47\textdegree & \fontsize{6.5}{6.5}\selectfont 0.061 & \fontsize{6.5}{6.5}\selectfont 0.081 & \fontsize{6.5}{6.5}\selectfont $NC$ & \fontsize{6.5}{6.5}\selectfont $NC$ & \fontsize{6.5}{6.5}\selectfont $NC$  \\
            \scriptsize IDAM$^{\dag}$({GNN})~\citep{li2019iterative}& \fontsize{6.5}{6.5}\selectfont 4.46\textdegree & \fontsize{6.5}{6.5}\selectfont 0.029 & \fontsize{6.5}{6.5}\selectfont 0.039 & \fontsize{6.5}{6.5}\selectfont 57.85\textdegree & \fontsize{6.5}{6.5}\selectfont 0.253 & \fontsize{6.5}{6.5}\selectfont 0.374 & 
            \fontsize{6.5}{6.5}\selectfont 4.25\textdegree & \fontsize{6.5}{6.5}\selectfont 0.025 & \fontsize{6.5}{6.5}\selectfont 0.037 & \fontsize{6.5}{6.5}\selectfont 48.23\textdegree & \fontsize{6.5}{6.5}\selectfont 0.182 & \fontsize{6.5}{6.5}\selectfont 0.333 \\
            \scriptsize IDAM$^{\dag}$(\textit{FPFH})~\citep{li2019iterative} & \fontsize{6.5}{6.5}\selectfont 9.60\textdegree & \fontsize{6.5}{6.5}\selectfont 0.052 & \fontsize{6.5}{6.5}\selectfont 0.084 & \fontsize{6.5}{6.5}\selectfont 71.06\textdegree & \fontsize{6.5}{6.5}\selectfont 0.217 & \fontsize{6.5}{6.5}\selectfont 0.430 & 
            \fontsize{6.5}{6.5}\selectfont 9.50\textdegree & \fontsize{6.5}{6.5}\selectfont 0.051 & \fontsize{6.5}{6.5}\selectfont 0.087 & \fontsize{6.5}{6.5}\selectfont 66.01\textdegree & \fontsize{6.5}{6.5}\selectfont 0.198 & \fontsize{6.5}{6.5}\selectfont 0.420  \\
            \scriptsize RGM$^{\dag}$~\citep{fu2021robust} & \fontsize{6.5}{6.5}\selectfont 2.21\textdegree & \fontsize{6.5}{6.5}\selectfont 0.013 & \fontsize{6.5}{6.5}\selectfont 0.018 & \fontsize{6.5}{6.5}\selectfont 23.58\textdegree & \fontsize{6.5}{6.5}\selectfont 0.111 & \fontsize{6.5}{6.5}\selectfont 0.156 & 
            \fontsize{6.5}{6.5}\selectfont 2.62\textdegree & \fontsize{6.5}{6.5}\selectfont 0.025 & \fontsize{6.5}{6.5}\selectfont 0.030 & \fontsize{6.5}{6.5}\selectfont 25.65\textdegree & \fontsize{6.5}{6.5}\selectfont 0.117 & \fontsize{6.5}{6.5}\selectfont 0.179  \\
            \scriptsize Predator$^{\dag}$~\citep{huang2021predator} & \fontsize{6.5}{6.5}\selectfont 3.33\textdegree & \fontsize{6.5}{6.5}\selectfont 0.018 & \fontsize{6.5}{6.5}\selectfont 0.026 & \fontsize{6.5}{6.5}\selectfont 40.64\textdegree & \fontsize{6.5}{6.5}\selectfont 0.110 & \fontsize{6.5}{6.5}\selectfont 0.207 & 
            \fontsize{6.5}{6.5}\selectfont 2.91\textdegree & \fontsize{6.5}{6.5}\selectfont 0.017 & \fontsize{6.5}{6.5}\selectfont 0.024 & \fontsize{6.5}{6.5}\selectfont 34.32\textdegree & \fontsize{6.5}{6.5}\selectfont \textbf{0.090} & \fontsize{6.5}{6.5}\selectfont 0.179 \\
            \scriptsize DCP~\citep{wang2019deep} & \fontsize{6.5}{6.5}\selectfont 9.33\textdegree & \fontsize{6.5}{6.5}\selectfont 0.070 & \fontsize{6.5}{6.5}\selectfont 0.097 & \fontsize{6.5}{6.5}\selectfont 73.61\textdegree & \fontsize{6.5}{6.5}\selectfont 0.185 & \fontsize{6.5}{6.5}\selectfont 0.441 &
            \fontsize{6.5}{6.5}\selectfont 10.58\textdegree & \fontsize{6.5}{6.5}\selectfont 0.072 & \fontsize{6.5}{6.5}\selectfont 0.109 & \fontsize{6.5}{6.5}\selectfont 75.34\textdegree & \fontsize{6.5}{6.5}\selectfont 0.192 & \fontsize{6.5}{6.5}\selectfont 0.467 \\
            \scriptsize DeepGMR (\textit{RRI})~\citep{yuan2020deepgmr} & \fontsize{6.5}{6.5}\selectfont 16.96\textdegree & \fontsize{6.5}{6.5}\selectfont 0.068 & \fontsize{6.5}{6.5}\selectfont 0.120 & \fontsize{6.5}{6.5}\selectfont 68.68\textdegree & \fontsize{6.5}{6.5}\selectfont 0.248 & \fontsize{6.5}{6.5}\selectfont 0.419 & 
            \fontsize{6.5}{6.5}\selectfont 17.67\textdegree & \fontsize{6.5}{6.5}\selectfont 0.067 & \fontsize{6.5}{6.5}\selectfont 0.131 & \fontsize{6.5}{6.5}\selectfont 41.60\textdegree & \fontsize{6.5}{6.5}\selectfont 0.153 & \fontsize{6.5}{6.5}\selectfont 0.308 \\
            \scriptsize DeepGMR (XYZ)~\citep{yuan2020deepgmr} & \fontsize{6.5}{6.5}\selectfont 6.48\textdegree & \fontsize{6.5}{6.5}\selectfont 0.049 & \fontsize{6.5}{6.5}\selectfont 0.064 & \fontsize{6.5}{6.5}\selectfont 70.26\textdegree & \fontsize{6.5}{6.5}\selectfont 0.246 & \fontsize{6.5}{6.5}\selectfont 0.428 & 
            \fontsize{6.5}{6.5}\selectfont 7.55\textdegree & \fontsize{6.5}{6.5}\selectfont 0.052 & \fontsize{6.5}{6.5}\selectfont 0.071 & \fontsize{6.5}{6.5}\selectfont 70.26\textdegree & \fontsize{6.5}{6.5}\selectfont 0.233 & \fontsize{6.5}{6.5}\selectfont 0.436 \\
            \scriptsize RPMNet (\textit{PPF})~\citep{yew2020rpm} & \fontsize{6.5}{6.5}\selectfont 3.52\textdegree & \fontsize{6.5}{6.5}\selectfont 0.214 & \fontsize{6.5}{6.5}\selectfont 0.029 & \fontsize{6.5}{6.5}\selectfont 37.82\textdegree & \fontsize{6.5}{6.5}\selectfont 0.132 & \fontsize{6.5}{6.5}\selectfont 0.250 &
            \fontsize{6.5}{6.5}\selectfont 3.80\textdegree & \fontsize{6.5}{6.5}\selectfont 0.022 & \fontsize{6.5}{6.5}\selectfont 0.032 & \fontsize{6.5}{6.5}\selectfont 39.52\textdegree & \fontsize{6.5}{6.5}\selectfont 0.129 & \fontsize{6.5}{6.5}\selectfont 0.271 \\
            \midrule[0.25pt]
            \scriptsize \OM{} (\textit{RRI})  & \fontsize{6.5}{6.5}\selectfont 1.28\textdegree & \fontsize{6.5}{6.5}\selectfont 0.009 & \fontsize{6.5}{6.5}\selectfont 0.011 & \fontsize{6.5}{6.5}\selectfont 50.62\textdegree & \fontsize{6.5}{6.5}\selectfont 0.206 & \fontsize{6.5}{6.5}\selectfont 0.315 & 
            \fontsize{6.5}{6.5}\selectfont 1.47\textdegree & \fontsize{6.5}{6.5}\selectfont 0.010 & \fontsize{6.5}{6.5}\selectfont 0.012 & \fontsize{6.5}{6.5}\selectfont 51.49\textdegree & \fontsize{6.5}{6.5}\selectfont 0.200 & \fontsize{6.5}{6.5}\selectfont 0.335 \\
            \scriptsize \OM{} (\textit{FPFH})  & \fontsize{6.5}{6.5}\selectfont 1.18\textdegree & \fontsize{6.5}{6.5}\selectfont 0.008 & \fontsize{6.5}{6.5}\selectfont 0.010 & \fontsize{6.5}{6.5}\selectfont 25.95\textdegree & \fontsize{6.5}{6.5}\selectfont 0.115 & \fontsize{6.5}{6.5}\selectfont 0.184 & 
            \fontsize{6.5}{6.5}\selectfont 1.33\textdegree & \fontsize{6.5}{6.5}\selectfont 0.009 & \fontsize{6.5}{6.5}\selectfont 0.011 & \fontsize{6.5}{6.5}\selectfont 33.21\textdegree & \fontsize{6.5}{6.5}\selectfont 0.140 & \fontsize{6.5}{6.5}\selectfont 0.242  \\
            \scriptsize \OM{} (\textit{PPF})  & \fontsize{6.5}{6.5}\selectfont \textbf{0.94\textdegree} & \fontsize{6.5}{6.5}\selectfont \textbf{0.007} & \fontsize{6.5}{6.5}\selectfont \textbf{0.008} &
            \fontsize{6.5}{6.5}\selectfont \textbf{18.13\textdegree} & \fontsize{6.5}{6.5}\selectfont \textbf{0.093} & \fontsize{6.5}{6.5}\selectfont \textbf{0.132} & 
            \fontsize{6.5}{6.5}\selectfont \textbf{1.12\textdegree} & \fontsize{6.5}{6.5}\selectfont \textbf{0.008} & \fontsize{6.5}{6.5}\selectfont \textbf{0.010} & \fontsize{6.5}{6.5}\selectfont \textbf{21.63\textdegree} & \fontsize{6.5}{6.5}\selectfont {0.111} & \fontsize{6.5}{6.5}\selectfont \textbf{0.169}  \\
            \bottomrule[1.2pt]
        \end{tabular}
    \end{center}
    \label{tab:mn40_noisy_and_unseen}
\end{table*}
\subsection{Registration on ModelNet40}
ModelNet40~\citep{wu20153d} contains CAD models of 40 categories,
which is split into 9,843 shapes for training and 2,468 for testing.
For each 3D shape, we uniformly sample 1,024 points from its surfaces.
To imitate partial scans, incomplete point clouds are generated by selecting the most nearest 768 points for a randomly placed point in the space~\citep{wang2019prnet}.
Following previous works~\citep{wang2019deep,yuan2020deepgmr,yew2020rpm}, 
we use four different data settings: 1) ``Clean'', 2) ``Unseen'', 3) ``Noisy'' and ``Noisy \& Unseen'', which are reported in 
Table~\ref{tab:mn40_clean_and_unseen} and Table~\ref{tab:mn40_noisy_and_unseen},
respectively.
Moreover, we evaluate both restricted rotations ``[0, 45\textdegree]'' and unrestricted rotations ``[0, 180\textdegree]'' for each data setting.
To achieve fair comparisons, we use the same rotation augmentations - that is [0, 45\textdegree] for restricted rotations and [0, 180\textdegree] for the unrestricted, for all reported methods. 
For all the other settings, we use their official configurations on ModelNet40, which are provided by the authors.
All models are trained with 200 epochs.
Note that many methods may use both handcrafted \textit{RI} features and spatial coordinate features $XYZ + \Delta XYZ$, although only different \textit{RI} features are highlighted. 

\noindent\textbf{Unseen Shape Point Cloud Registration.}
We use point clouds from all the 40 categories of ModelNet40 for training and testing.
Note that all the objects used for testing have not been seen during training, and therefore in this setting we evaluate the PPR for unseen 3D shapes.
As reported in 
Table~\ref{tab:mn40_clean_and_unseen},
\OM{} using different \textit{RI} features, \textit{RRI}, \textit{FPFH} and \textit{PPF}, achieves nearly perfect registration results, even for full-range (\emph{i.e,} [0, 180\textdegree]) PPR, which challenges previous SoTA methods.
Qualitative results are shown in Fig.~\ref{fig:mn40}, and \OM{} could estimate correct pose-invariant correspondences for registration.

\noindent\textbf{Unseen Category Point Cloud Registration.}
In this setting, we train our network by using 3D point clouds from the first 20 categories, and use the other 20 category point clouds for evaluation.
Because \OM{} learns to encode local transformation-robust geometrical features that do not rely on the global shape category information heavily, it shows great generalizability for registering point clouds from unseen categories.
Similar with the ``Clean'' setting, \OM{} significantly outperforms existing SoTA methods in 
Table~\ref{tab:mn40_clean_and_unseen},
especially for ``Unseen'' full-range PPR.

\begin{table*}
\setlength{\tabcolsep}{8pt}
    \caption{\textbf{Input Feature Analysis.}
    {Handcrafted rotation-invariant features (\textit{RRI}, \textit{FPFH}, and \textit{PPF}) and ``XYZ+$\Delta$XYZ'' features.}
    $^\ddag$ denotes the issue ``svd did not converge'' occurred.
    }
    \begin{center}
        \begin{tabular}{c|c|ccc|ccc|ccc}
            \toprule[1.2pt]
            \multirow{2}{*}{\makecell{ \fontsize{8.5}{8.5}\selectfont Input \\ \fontsize{8.5}{8.5}\selectfont Features }} & 
            \multirow{2}{*}{\makecell{\fontsize{8.5}{8.5}\selectfont Rotation \\ \fontsize{8.5}{8.5}\selectfont Ranges }} & 
            \multicolumn{3}{c|}{\fontsize{9}{9}\selectfont Clean} & \multicolumn{3}{c|}{\fontsize{9}{9}\selectfont Unseen} & \multicolumn{3}{c}{\fontsize{9}{9}\selectfont Noise} \\
            \cmidrule[0.25pt]{3-11}
            & & \fontsize{8.5}{8.5} $\mathcal{L}_{\mathbf{R}}$ & \fontsize{8.5}{8.5}\selectfont $\mathcal{L}_{\mathbf{t}}$ & {\fontsize{8.5}{8.5}\selectfont $\mathcal{L}_{\text{RMSE}}$} & \fontsize{8.5}{8.5}\selectfont $\mathcal{L}_{\mathbf{R}}$ & \fontsize{8.5}{8.5}\selectfont $\mathcal{L}_{\mathbf{t}}$ & {\fontsize{8.5}{8.5}\selectfont $\mathcal{L}_{\text{RMSE}}$} & \fontsize{8.5}{8.5}\selectfont $\mathcal{L}_{\mathbf{R}}$ & \fontsize{8.5}{8.5}\selectfont $\mathcal{L}_{\mathbf{t}}$ & {\fontsize{8.5}{8.5}\selectfont $\mathcal{L}_{\text{RMSE}}$}
            \\
            \midrule[0.75pt]
            \multirow{2}{*}{\fontsize{8.5}{8.5}\selectfont \textit{RRI} Only} & \fontsize{8.5}{8.5}\selectfont [0, 45\text{\textdegree}] & 
            \fontsize{7.5}{7.5}\selectfont 0.015\textdegree & \fontsize{7.5}{7.5}\selectfont \textless 0.0001 & \fontsize{7.5}{7.5}\selectfont \textless 0.0001 & \fontsize{7.5}{7.5}\selectfont 0.023\textdegree & \fontsize{7.5}{7.5}\selectfont 0.0002 & \fontsize{7.5}{7.5}\selectfont 0.0002 & \fontsize{7.5}{7.5}\selectfont 10.50\textdegree & \fontsize{7.5}{7.5}\selectfont 0.050 & \fontsize{7.5}{7.5}\selectfont 0.075 \\
             & \fontsize{8.5}{8.5}\selectfont [0, 180\text{\textdegree}] &
            \fontsize{7.5}{7.5}\selectfont \textbf{0.021\textdegree} & \fontsize{7.5}{7.5}\selectfont \textbf{0.0001} & \fontsize{7.5}{7.5}\selectfont \textbf{0.0001} & \fontsize{7.5}{7.5}\selectfont \textbf{0.043\textdegree} & \fontsize{7.5}{7.5}\selectfont \textbf{0.0003} & \fontsize{7.5}{7.5}\selectfont \textbf{0.0003} & \fontsize{7.5}{7.5}\selectfont \textbf{49.12\textdegree} & \fontsize{7.5}{7.5}\selectfont \textbf{0.201} & \fontsize{7.5}{7.5}\selectfont \textbf{0.306} \\
            \midrule[0.25pt]
            \multirow{2}{*}{\makecell{\fontsize{8.5}{8.5}\selectfont \textit{RRI} \\ \fontsize{8.5}{8.5}\selectfont + XYZ + $\Delta$XYZ }} & \fontsize{8.5}{8.5}\selectfont [0, 45\text{\textdegree}] & 
             \fontsize{7.5}{7.5}\selectfont \textbf{0.013\textdegree} & \fontsize{7.5}{7.5}\selectfont \textbf{\textless 0.0001} & \fontsize{7.5}{7.5}\selectfont \textbf{\textless 0.0001} & \fontsize{7.5}{7.5}\selectfont \textbf{0.017\textdegree} & \fontsize{7.5}{7.5}\selectfont \textbf{\textless 0.0001} & \fontsize{7.5}{7.5}\selectfont \textbf{\textless 0.0001} & \fontsize{7.5}{7.5}\selectfont \textbf{1.28\textdegree} & \fontsize{7.5}{7.5}\selectfont \textbf{0.009} & \fontsize{7.5}{7.5}\selectfont \textbf{0.011} \\
             & \fontsize{8.5}{8.5}\selectfont [0, 180\text{\textdegree}] & \fontsize{7.5}{7.5}\selectfont 0.036\textdegree & \fontsize{7.5}{7.5}\selectfont 0.0002 & \fontsize{7.5}{7.5}\selectfont 0.0002 & \fontsize{7.5}{7.5}\selectfont 0.053\textdegree & \fontsize{7.5}{7.5}\selectfont 0.0004 & \fontsize{7.5}{7.5}\selectfont 0.0005 & \fontsize{7.5}{7.5}\selectfont {50.62}\textdegree & \fontsize{7.5}{7.5}\selectfont {0.206} & \fontsize{7.5}{7.5}\selectfont {0.315} \\
            \midrule[0.75pt]
             \multirow{2}{*}{\fontsize{8.5}{8.5}\selectfont \textit{FPFH} Only} & \fontsize{8.5}{8.5}\selectfont [0, 45\text{\textdegree}] & 
            \fontsize{7.5}{7.5}\selectfont 0.206\textdegree & \fontsize{7.5}{7.5}\selectfont 0.0024 & \fontsize{7.5}{7.5}\selectfont 0.0025 & \fontsize{7.5}{7.5}\selectfont 0.387\textdegree & \fontsize{7.5}{7.5}\selectfont 0.0038 & \fontsize{7.5}{7.5}\selectfont 0.0041 & \fontsize{7.5}{7.5}\selectfont 7.97\textdegree & \fontsize{7.5}{7.5}\selectfont 0.045 & \fontsize{7.5}{7.5}\selectfont 0.065 \\
             & \fontsize{8.5}{8.5}\selectfont [0, 180\text{\textdegree}] &
            \fontsize{7.5}{7.5}\selectfont {2.43\textdegree}$^{\ddag}$ & \fontsize{7.5}{7.5}\selectfont {0.0203}$^{\ddag}$ & \fontsize{7.5}{7.5}\selectfont {0.0223}$^{\ddag}$ & \fontsize{7.5}{7.5}\selectfont {2.33\textdegree}$^{\ddag}$ & \fontsize{7.5}{7.5}\selectfont {0.0200}$^{\ddag}$ & \fontsize{7.5}{7.5}\selectfont {0.0226}$^{\ddag}$ & \fontsize{7.5}{7.5}\selectfont 29.93\textdegree & \fontsize{7.5}{7.5}\selectfont 0.130 & \fontsize{7.5}{7.5}\selectfont 0.210 \\
            \midrule[0.25pt]
            \multirow{2}{*}{\makecell{\fontsize{8.5}{8.5}\selectfont \textit{FPFH} \\ \fontsize{8.5}{8.5}\selectfont + XYZ + $\Delta$XYZ }} & \fontsize{8.5}{8.5}\selectfont [0, 45\text{\textdegree}] & 
             \fontsize{7.5}{7.5}\selectfont \textbf{0.012\textdegree} & \fontsize{7.5}{7.5}\selectfont \textbf{\textless 0.0001} & \fontsize{7.5}{7.5}\selectfont \textbf{\textless 0.0001} & \fontsize{7.5}{7.5}\selectfont \textbf{0.014\textdegree} & \fontsize{7.5}{7.5}\selectfont \textbf{\textless 0.0001} & \fontsize{7.5}{7.5}\selectfont \textbf{\textless 0.0001} & \fontsize{7.5}{7.5}\selectfont \textbf{1.18\textdegree} & \fontsize{7.5}{7.5}\selectfont \textbf{0.008} & \fontsize{7.5}{7.5}\selectfont \textbf{0.010} \\
             & \fontsize{8.5}{8.5}\selectfont [0, 180\text{\textdegree}] & \fontsize{7.5}{7.5}\selectfont \textbf{0.253\textdegree} & \fontsize{7.5}{7.5}\selectfont \textbf{0.0008} & \fontsize{7.5}{7.5}\selectfont \textbf{0.0016} & \fontsize{7.5}{7.5}\selectfont \textbf{0.549\textdegree} & \fontsize{7.5}{7.5}\selectfont \textbf{0.0024} & \fontsize{7.5}{7.5}\selectfont \textbf{0.0033} & \fontsize{7.5}{7.5}\selectfont \textbf{25.95}\textdegree & \fontsize{7.5}{7.5}\selectfont \textbf{0.115} & \fontsize{7.5}{7.5}\selectfont \textbf{0.184} \\
            \midrule[0.75pt]
            \multirow{2}{*}{\fontsize{8.5}{8.5}\selectfont \textit{PPF} Only} & \fontsize{8.5}{8.5}\selectfont [0, 45\text{\textdegree}] & 
            \fontsize{7.5}{7.5}\selectfont 0.017\textdegree & \fontsize{7.5}{7.5}\selectfont 0.0002 & \fontsize{7.5}{7.5}\selectfont 0.0002 & \fontsize{7.5}{7.5}\selectfont 0.039\textdegree & \fontsize{7.5}{7.5}\selectfont 0.0005 & \fontsize{7.5}{7.5}\selectfont 0.0005 & \fontsize{7.5}{7.5}\selectfont 4.41\textdegree & \fontsize{7.5}{7.5}\selectfont 0.030 & \fontsize{7.5}{7.5}\selectfont 0.038 \\
             & \fontsize{8.5}{8.5}\selectfont [0, 180\text{\textdegree}] &
            \fontsize{7.5}{7.5}\selectfont \textbf{0.258\textdegree} & \fontsize{7.5}{7.5}\selectfont \textbf{0.0020} & \fontsize{7.5}{7.5}\selectfont \textbf{0.0023} & \fontsize{7.5}{7.5}\selectfont \textbf{0.238\textdegree} & \fontsize{7.5}{7.5}\selectfont \textbf{0.0017} & \fontsize{7.5}{7.5}\selectfont \textbf{0.0018} & \fontsize{7.5}{7.5}\selectfont 23.13\textdegree & \fontsize{7.5}{7.5}\selectfont 0.116 & \fontsize{7.5}{7.5}\selectfont 0.168 \\
            \midrule[0.25pt]
            \multirow{2}{*}{\makecell{\fontsize{8.5}{8.5}\selectfont \textit{PPF} \\ \fontsize{8.5}{8.5}\selectfont + XYZ + $\Delta$XYZ }} & \fontsize{8.5}{8.5}\selectfont [0, 45\text{\textdegree}] & 
             \fontsize{7.5}{7.5}\selectfont \textbf{0.016\textdegree} & \fontsize{7.5}{7.5}\selectfont \textbf{\textless 0.0001} & \fontsize{7.5}{7.5}\selectfont \textbf{\textless 0.0001} & \fontsize{7.5}{7.5}\selectfont \textbf{0.026\textdegree} & \fontsize{7.5}{7.5}\selectfont \textbf{0.0002} & \fontsize{7.5}{7.5}\selectfont \textbf{0.0002} & \fontsize{7.5}{7.5}\selectfont \textbf{0.94\textdegree} & \fontsize{7.5}{7.5}\selectfont \textbf{0.007} & \fontsize{7.5}{7.5}\selectfont \textbf{0.008} \\
             & \fontsize{8.5}{8.5}\selectfont [0, 180\text{\textdegree}] & \fontsize{7.5}{7.5}\selectfont 0.284\textdegree & \fontsize{7.5}{7.5}\selectfont 0.0011 & \fontsize{7.5}{7.5}\selectfont 0.0017 & \fontsize{7.5}{7.5}\selectfont 0.388\textdegree & \fontsize{7.5}{7.5}\selectfont 0.0024 & \fontsize{7.5}{7.5}\selectfont 0.0030 & \fontsize{7.5}{7.5}\selectfont \textbf{18.13}\textdegree & \fontsize{7.5}{7.5}\selectfont \textbf{0.093} & \fontsize{7.5}{7.5}\selectfont \textbf{0.132} \\
            \bottomrule[1.2pt]
        \end{tabular}
    \end{center}
    \label{tab:feats}
\end{table*}

\noindent\textbf{Noisy Point Cloud Registration.}
Following previous methods, we add Gaussian noise that are randomly sampled from $\mathcal{N}(0, \, 0.01)$ and clipped to $[-0.05, \, 0.05]$ to partial point clouds for evaluating PPR under random noise.
By adding random noise, raw handcrafted \textit{RI} features become no longer transformation-invariant, which mostly influences the global registration performance.
In our experiments reported in 
Table~\ref{tab:mn40_noisy_and_unseen},
\OM{} could resolve local PPR with small registration error, and also achieves better global registration with the help of \textit{PPF} than previous SoTA methods.
Because we are focused on encoding geometrical features from local structures of the 3D shapes, those symmetric shapes and repetitive structures (\emph{e.g,} vases and tables) of 3D shapes from ModelNet40 dataset give rise to challenges, such as plausible feature correspondences or local optima, in resolving noisy full-range PPR.

\noindent\textbf{Unseen Category and Noisy Point Cloud Registration.}
The ``Unseen \& Noisy'' setting is a combination of ``Unseen'' and ``Noisy'': 
Firstly, we train networks with 3D objects from the first 20 categories, and then test on the rest 20 categories;
additionally, random noise sampled from Gaussian distribution $\mathcal{N}(0, \, 0.01)$ that is clipped to $[-0.05, \, 0.05]$, is added to the partial point clouds.
Similar with the ```Noisy'' setting, the proposed \OM{} could resolve local registration very well, but also struggles with global registration.
Even though, \OM{} achieves better registration than previous SoTA methods for most evaluated metrics.

\begin{figure*}
    \centering
    \includegraphics[width=1\linewidth]{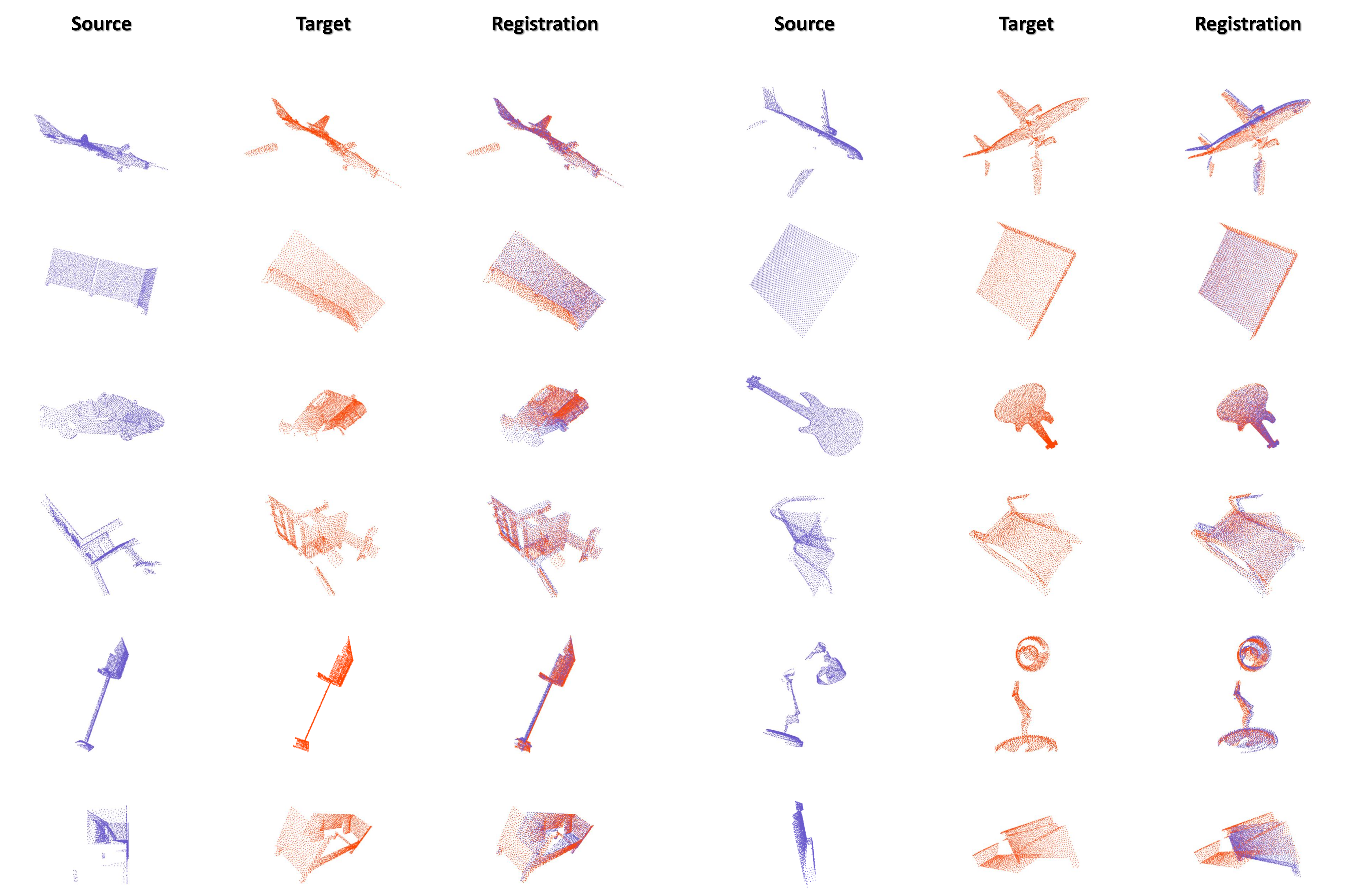}
    \caption{
    Partial-to-Partial Point Cloud Registration Results by \OM{} on MVP-RG Dataset.}
    \label{fig:mvp_rg}
\end{figure*}
\noindent\textbf{Different Rotation-Invariant ({\textit{RI}}) Features.}
In Table~\ref{tab:mn40_clean_and_unseen} and Table~\ref{tab:mn40_noisy_and_unseen}, we evaluated \textit{RRI}, \textit{FPFH} and \textit{PPF} in \OM{}.
Although \textit{RRI} features are no longer consistent for clean partial point clouds under different SE(3) transformations due to center shifting (as discussed in Sec.~\ref{sec:pa}), \OM{} estimate faithful correspondences between different partial point clouds for PPR. 
We think the reason is that the inconsistency caused by translations influences the encoded \textit{RRI} features for all points, which is compensated during the feature matching optimization if perfect correspondences exist. 
Together with random noise, local registration with \textit{RRI} features can be resolved mainly due to using spatial coordinates graphs.
However, it is too challenging to use \textit{RRI} for global PPR with noisy data, while those local \textit{RI} features, \textit{FPFH} and \textit{PPF}, can lead to better registration results. 

\begin{table}
\setlength{\tabcolsep}{9pt}
    \caption{\textbf{Ablation studies} for full-range Partial-to-Partial Point Cloud Registration on ModelNet40 (``Clean''). 
    }
    \vspace{-2mm}
    \begin{center}
    \small{
        \begin{tabular}{ccc|ccc}
            \toprule[1.2pt]
            \scriptsize K-NN & \scriptsize TPT & \scriptsize HGM & \scriptsize $\mathcal{L}_{\mathbf{R}}\downarrow$ & \scriptsize $\mathcal{L}_{\mathbf{t}}\downarrow$ & \multicolumn{1}{c}{\scriptsize $\mathcal{L}_{\text{RMSE}}\downarrow$} \\
            \midrule[0.75pt]
            & & & \fontsize{7.5}{7.5}\selectfont 36.93\textdegree & \fontsize{7.5}{7.5}\selectfont 0.197 & \fontsize{7.5}{7.5}\selectfont 0.286 \\
            \checkmark & & & \fontsize{7.5}{7.5}\selectfont 9.01\textdegree & \fontsize{7.5}{7.5}\selectfont 0.048 & \fontsize{7.5}{7.5}\selectfont 0.065 \\
            \checkmark & \checkmark & & \fontsize{7.5}{7.5}\selectfont 5.85\textdegree & \fontsize{7.5}{7.5}\selectfont 0.036 & \fontsize{7.5}{7.5}\selectfont 0.065 \\
            \checkmark & \checkmark & \checkmark & \fontsize{7.5}{7.5}\selectfont 0.016\textdegree & \fontsize{7.5}{7.5}\selectfont \textless 0.0001 & \fontsize{7.5}{7.5}\selectfont \textless 0.0001 \\
            \bottomrule[1.2pt]
        \end{tabular}
    }
    \end{center}
    \label{tab:ablation}
\end{table}
\noindent\textbf{Effectiveness of Different Input Features.}
We evaluate the effectiveness of different input feature elements, including \textit{RRI}, \textit{FPFH}, \textit{PPF}, XYZ and $\Delta$XYZ, in Table~\ref{tab:feats}.
Obviously, all handcrafted \textit{RI} features can resolve full-range registration with clean data (``Clean'' and ``Unseen''), and using \textit{RRI} features achieves slightly better performance than using \textit{PPF} or \textit{FPFH}.
We think it is because the center shifting problem of \textit{RRI} features can be compensated during optimization, while surface normal direction ambiguities cannot.
However, they perform differently with noisy point clouds (``Noisy'').
With small relative transformations (\emph{i.e.} [0, 45\textdegree{}]), using spatial coordinate features $XYZ + \Delta XYZ$ significantly improves PPR performance for all handcrafted \textit{RI} features.
For large relative transformations (\emph{i.e.} [0, 180\textdegree{}]), the PPR problem is too challenging to be fully resolved.
The global \textit{RI} feature, \textit{RRI}, performs the worst on noisy point clouds than the other local \textit{RI} features, \textit{FPFH} and \textit{PPF}.
Comparing to \textit{FPFH}, \textit{PPF} is preferred due to 1) it achieves better performance; and 2) it is more convenient to construct in terms of time efficiency.
We think the main challenges for global PPR with noisy data are that many objects in ModelNet40 and MVP-RG
1) are rotational symmetric;
or 2) have repetitive similar structures.
Hence, if there are few or no perfect correspondences caused by noise, PPR can often converge to other alignment results rather the groundtruth.
For image feature matching, recent methods, such as R2D2~\citep{revaud2019r2d2} and SuperGlue~\citep{sarlin2020superglue}, are proposed, which can be a promising future research direction for 3D feature matching.

\noindent\textbf{Ablation Study.}
We report ablation studies on the main components of \OM{}, 
including using K-NN graphs, TPT modules 
and HGM architecture, in Table~\ref{tab:ablation}.
The baseline model directly uses \textit{PPF} features for registration.
With the help of K-NN graphs, the encoded features become more discriminative, and hence leading to much better registration.
Using TPT modules further improves registration performance.
Moreover, 
HGM introduces large-scale graph-based relations based on TPT and the hierarchical architecture, which perfectly resolves most PPR with clean data.
In case that readers are curious about the effectiveness of using \textit{RI} features, we further report two ablation settings,
\textbf{1) \textit{RI} features + RANSAC:} we try to match \textit{PPF} by using RANSAC with 4 million iterations by tuning different inlier ratio settings. 
The better results of RANSAC(4M) are 14.69\textdegree ($\mathcal{L}_{\mathbf{R}}$), 0.149 ($\mathcal{L}_{\mathbf{t}}$), and 0.159 ($\mathcal{L}_{\text{RMSE}}$) on the ModelNet40 with the ``Clean'' settings.
Because RANSAC cannot adapt to different inlier ratios, registration by RANSAC(4M) is much worse than \OM{}.
\textbf{2) No \textit{RI} features:} As they are the key components of our methods, without using \textit{RI} features, OM{} could not converge under large transformations.  
Note that \OM{} does not explicitly supervise the corresponding feature points to have similar feature descriptors during training, but \textit{RI} features, especially local \textit{RI} features, could provide the geometrical clues. 

\subsection{Registration on MVP-RG}
We further evaluate \OM{} on the MVP-RG dataset consisting of  
view-based partial point clouds (see Fig.~\ref{fig:mvp}). 
Note that the partial point cloud pairs can have no perfect correspondences, since different partial point clouds are generated by different scans from various camera views. 
Hence, registering partial point clouds from MVP-RG dataset is a noisy full-range PPR problem.
To achieve fair comparisons, we use the same rotation augmentations for all methods, and use their best settings that previously are used for PPR with noisy data from ModelNet40.
All models are trained with 200 epochs.
Similarly, it is very challenging to register noisy partial point clouds for MVP-RG.
As reported in Table~\ref{tab:mvp}, \OM{} still outperforms previous SoTA methods on MVP-RG.
Qualitative registration results are shown in Fig.~\ref{fig:mvp_rg}.
\begin{table}[]
    \setlength{\tabcolsep}{10pt}
    \caption{
    \textbf{Partial-to-Partial Point Cloud Registration on MVP-RG.} 
    }
    \vspace{-2mm}
    \begin{center}
        \small{
        \begin{tabular}{l|ccc}
            \toprule[1.2pt]
             & \scriptsize $\mathcal{L}_{\mathbf{R}}$ & \scriptsize $\mathcal{L}_{\mathbf{t}}$ & \scriptsize $\mathcal{L}_{\text{RMSE}}$ \\
            \midrule[0.75pt]
            \scriptsize IDAM$^{\dag}$(\text{GNN})~\citep{li2019iterative}& \fontsize{7.5}{7.5}\selectfont 24.35\textdegree & \fontsize{7.5}{7.5}\selectfont 0.280 & \fontsize{7.5}{7.5}\selectfont 0.344 \\
            \scriptsize IDAM$^{\dag}$(\text{FPFH})~\citep{li2019iterative}& \fontsize{7.5}{7.5}\selectfont 35.78\textdegree & \fontsize{7.5}{7.5}\selectfont 0.391 & \fontsize{7.5}{7.5}\selectfont 0.476 \\
            \scriptsize RGM$^{\dag}$~\citep{fu2021robust} & \fontsize{7.5}{7.5}\selectfont 41.27\textdegree & \fontsize{7.5}{7.5}\selectfont 0.425 & \fontsize{7.5}{7.5}\selectfont 0.583 \\
            \scriptsize DCP~\citep{wang2019deep} & \fontsize{7.5}{7.5}\selectfont 30.37\textdegree & \fontsize{7.5}{7.5}\selectfont 0.273 & \fontsize{7.5}{7.5}\selectfont 0.634 \\
            \scriptsize DeepGMR (\textit{RRI})~\citep{yuan2020deepgmr} & \fontsize{7.5}{7.5}\selectfont 49.72\textdegree & \fontsize{7.5}{7.5}\selectfont 0.385 & \fontsize{7.5}{7.5}\selectfont 0.696 \\
            \scriptsize DeepGMR (XYZ)~\citep{yuan2020deepgmr} & \fontsize{7.5}{7.5}\selectfont 43.74\textdegree & \fontsize{7.5}{7.5}\selectfont 0.353 & \fontsize{7.5}{7.5}\selectfont 0.608 \\
            \scriptsize RPMNet (\textit{PPF})~\citep{yew2020rpm} & \fontsize{7.5}{7.5}\selectfont 22.20\textdegree & \fontsize{7.5}{7.5}\selectfont 0.174 & \fontsize{7.5}{7.5}\selectfont 0.327 \\
            \midrule[0.25pt]
            \scriptsize \OM{} (\textit{PPF})  & \fontsize{7.5}{7.5}\selectfont \textbf{16.57\textdegree} & \fontsize{7.5}{7.5}\selectfont \textbf{0.174} & \fontsize{7.5}{7.5}\selectfont \textbf{0.246}  \\
            \bottomrule[1.2pt]
        \end{tabular}
        }
    \end{center}
    
    \label{tab:mvp}
\end{table}

\section{Conclusion and Discussion}
In this work, we propose a comprehensive paradigm \OM{}, which utilizes a synergy of hierarchical graph networks and graphical modeling for object-centric PPR.
In particular, we provide an in-depth analysis on the rotation-invariance and noise-resilience for handcrafted features, based on which we propose a novel TPT model to adaptively aggregate neighboring features. 
Furthermore, we establish a challenging virtual-scanned partial point cloud dataset MVP-RG.
Extensive experiments show that \OM{} outperforms previous SoTA methods.

We want to highlight that in line with previous settings that evaluating object-centric PCR problem under local transformations ([0, 45\textdegree]), our \OM{} achieves much better registration results and almost perfectly resolve all the challenging PPR problems.
Furthermore, we evaluate on global registration problem ([0, 180\textdegree]), and the proposed \OM{} addresses most PPR problems excepting for registering those noisy partial point clouds (\emph{e.g.} ModelNet40(Noisy) and MVP-RG).
Even though, \OM{} still outperforms previous methods.
For future research, we encourage that researchers can learn to detect discriminative feature points by considering cross-contextual geometric information between the paired point clouds, analogy to image feature matching counterparts~\citep{revaud2019r2d2,sarlin2020superglue}.
We hope our study on learning transformation-robust point descriptors from handcrafted \textit{RI} features and the established MVP-RG dataset can inspire future research for 3D point cloud registration.

\section*{Acknowledgments}
This work is supported by NTU NAP, MOE AcRF Tier 2 (T2EP20221-0033), and under the RIE2020 Industry Alignment Fund – Industry Collaboration Projects (IAF-ICP) Funding Initiative, as well as cash and in-kind contribution from the industry partner(s).

\bibliographystyle{spbasic} 
\bibliography{egbib}

\end{document}